\documentclass{main}

\usepackage{graphicx}

\usepackage{subcaption}

\usepackage{amsmath}
\usepackage{amsfonts}

\usepackage{hyperref} 
\hypersetup{pdfborder=0 0 0}

\usepackage{doi}

\usepackage[square,numbers,sort&compress]{natbib}
\newcommand{\eqnref}[1]{Eq.(\ref{#1})}
\newcommand{\figref}[1]{Fig.\ref{#1}}
\newcommand{\tabref}[1]{Tab.\ref{#1}}

\usepackage{bm}
\usepackage{comment}
\usepackage{mathrsfs}   

\newcommand{\x}{{\bm{x}}}

\newcommand{\y}{{\bm{y}}}
\newcommand{\f}{{\bm{f}}}

\newcommand{\hyi}{\hat{y}_i(\x)}
\newcommand{\hsi}{s_i^2(\x)}

\usepackage[ruled,vlined]{algorithm2e}  
\usepackage{changepage}   

\title{Multi-objective Bayesian optimization with mixed-categorical design variables for expensive-to-evaluate aeronautical applications}

\author{
N. Bartoli$^{\bf {1*}}$, T. Lefebvre$^{\bf {1}}$,  R. Lafage$^{\bf {1}}$, P. Saves$^{\bf {1,2}}$, Y. Diouane$^{\bf {3}}$, J. Morlier$^{\bf {2}}$, J. H. Bussemaker$^{\bf {4}}$, G. Donelli$^{\bf {4}}$, J. M. Gomes de Mello$^{\bf {5}}$, M. Mandorino$^{\bf {6}}$, P. Della Vecchia$^{\bf {6}}$}
\heading{N. Bartoli, T. Lefebvre, R. Lafage, P. Saves, Y. Diouane, J. Morlier, J.H. Bussemaker, G. Donelli, Joao Marcos Gomes de Mello, M. Mandorino and P. Della Vecchia }

\address{1: ONERA/DTIS, \\ Université de Toulouse, \\ 31400 Toulouse, France \\
\{nathalie.bartoli,thierry.lefebvre,remi.lafage,paul.saves\}@onera.fr\\
\
\\
2: ICA,  ISAE--SUPAERO, INSA, CNRS, MINES ALBI, UPS\\
Université de Toulouse, \\ 31400 Toulouse, France \\    
joseph.morlier@isae-supaero.fr \\
\
\\
3: Polytechnique Montréal\\
Montreal, QC, Canada\\
youssef.diouane@polymtl.ca\\
\
\\
4: German Aerospace Center (DLR)\\
Institute of System Architectures in Aeronautics\\
Hein-Saß-Weg 22, 21129 Hamburg, Germany\\
\{jasper.bussemaker,giuseppa.donelli\}@dlr.de\\
\
\\
5: EMBRAER S.A.\\
Av. Brigadeiro Faria Lima, 2170, São José dos Campos, Brazil\\
joao.mello@embraer.com.br\\
\
\\
6: UNINA\\
University di Napoli Federico II\\
Naples, 80125, Italy\\
\{massimo.mandorino,pierluigi.dellavecchia\}@unina.it}

\abstract{This work aims at developing new methodologies to optimize computational costly complex systems (e.g., aeronautical engineering systems). The proposed surrogate-based method (often called Bayesian optimization) uses adaptive sampling to promote a trade-off between exploration and exploitation. Our in-house implementation, called SEGOMOE, handles a high number of design variables (continuous, discrete or categorical) and nonlinearities by combining mixtures of experts for the objective and/or the constraints. Additionally, the method handles multi-objective optimization settings, as it allows the construction of accurate Pareto fronts with a minimal number of function evaluations. Different infill criteria have been implemented to handle multiple objectives with or without constraints.
The effectiveness of the proposed method was tested on practical aeronautical applications within the context of the European Project AGILE 4.0 and demonstrated favorable results. A first example concerns a retrofitting problem where a comparison between two optimizers have been made. A second example introduces hierarchical variables to deal with architecture system in order to design an aircraft family. The third example increases drastically the number of categorical variables as it combines aircraft design, supply chain and manufacturing process. In this article, we show, on three different realistic problems, various aspects of our optimization codes thanks to the diversity of the treated aircraft problems.
}


\begin{document}

%

\section{Introduction}
For the last few decades, the aeronautical industry has been essentially driven by incremental improvements based on aircraft design optimization~\cite{chan2022trying}. The use of computer science allowed to reuse well-proven design processes and to benefit from preexisting results to reduce the time dedicated to development. Nonetheless, the forthcoming challenges awaiting aircraft design call for large-scale improvements to meet future expectations in terms of overall environmental impact, noise reduction and cost-effectiveness~\cite{Fioriti2022}. These large-scale improvements, considered through novel aircraft configurations, involve an increased proximity between Aerodynamics, Propulsion and Structural mechanics, thus requiring to explore coupled and uncharted physics.
As is the case for many industrial complex systems, some  powerful optimization algorithms for aircraft design are needed.
The H2020 European project AGILE 4.0 (2019-2023)\footnote{\url{https://www.agile4.eu/}}~\cite{agile4is2023}, led by DLR, has identified several challenging application cases linked to Multidisciplinary Design Optimization (MDO) methods~\cite{Martins:AIAAJ:2013} and Model Based Systems Engineering (MBSE) technologies~\cite{bussemaker2022ac7} that will require innovative and sustainable aircraft configurations in the years to come~\cite{AGILE4_INCOSE_1:23}. The main objective is to reduce the environmental impact in terms of fuel consumption, waste and emissions associated with all the aeronautical system activities and operations. This purpose requires extending aeronautical research to cover the entire aircraft lifecycle, from design and production to waste disposal after the system's end-of-life. Consequently, the challenge is to incorporate these new requirements into the early design phase, enabling strategic decision-making processes to optimize the entire aircraft lifecycle~\cite{chan2022aircraft}. To do this, collaborative multidisciplinary aircraft design and optimization should be carried out involving not-only the aircraft design domain (typically accounted during the conceptual aircraft design) but also other industrial domains, such as manufacturing, supply chain, maintenance or certification~\cite{AGILE4_INCOSE_2:23}. 
From the different application cases of the project, some quantities of interest have been identified from the Multidisciplinary Design Analysis (MDA) and optimization problems have been deduced involving from 2 to 5 objective functions and several nonlinear constraints that are given by non derivative and expensive-to-evaluate black box functions. Moreover, when evaluating new architecture systems, it is important to take into account categorical variables such as on-board system architecture (conventional, hybrid, electric for instance) or property materials (aluminum, titanium), which do not have a defined order~\cite{Mixed_Paul}. 
Different approaches have been investigated to deal with mixed integer variables, multiple objectives and constraints handling requirements. These optimizations return a set of feasible configurations considered as optimal trade-offs. this set of solutions is the so-called \textit{Pareto front} that consists in the set of solutions that cannot be improved in one objective without worsening another~\cite{savesMOO_AIAA:22}. 

Surrogate-Based Optimization (SBO) is a powerful technique for optimizing complex systems requiring a significant amount of computational resources. For multi-objective optimization problems, it involves building a surrogate model, such as a machine learning model, to approximate the behavior of multiple objectives simultaneously, allowing for efficient exploration of the search space and identification of the Pareto front. The surrogate models built offline are then used to predict the objective function values at different points in the search space, enabling the optimization algorithm to focus its computational resources on the most promising regions of the search space~\cite{jones1998efficient}. 

To avoid this offline cost involving a large number of function evaluations, some adaptive strategies have been introduced. Nowadays, Bayesian Optimization (BO)~\cite{garnett_bayesoptbook_2023} is well known as a powerful tool in scientific research for optimizing complex, expensive-to-evaluate functions with unknown properties. It is particularly useful when only a limited number of evaluations is available which is often the case with scientific experiments.
When multiple objectives are involved, multi-objective techniques must be used  to find the set of optimal solutions that maximize or minimize these multiple objectives simultaneously, rather than focusing on a single objective. Multi-objective optimization using Bayesian algorithms is a potent strategy for optimizing intricate systems that possess multiple, conflicting objectives. Bayesian optimization techniques allow for efficient exploration of the search space by harnessing probabilistic models to forecast the behavior of the objectives being optimized. This facilitates a trade-off between exploration and exploitation and allows the inclusion of prior knowledge on the objectives.

In summary, multi-objective optimization with Bayesian algorithms is a valuable tool for scientific research, providing an efficient way to identify the Pareto front and make informed decisions based on multiple, conflicting objectives. Here some illustrations are done on three application cases of the AGILE 4.0 project with different characteristics. On the first application case, about retrofitting concepts (engine, winglet, on-board systems), the different impacts on aircraft performance are characterized (e.g., Direct Operating Cost (DOC), emissions) to select the best retrofitting strategy. The second application is about designing an aircraft family composed of three different aircrafts with different commonality choices (wing, engine, empennage). Again, the different solutions are characterized by different impacts on aircraft performance (e.g. mass, aerodynamic efficiency) and different commonality choices. To finish with, the third application concerns Horizontal Tail Plane (HTP) production and search for a equilibrium between airplane performance, manufacturing costs and the efficiency of the supply chain.

The outline of the paper is as follows.
In Section~\ref{sec:multiobj}, the multi-objective optimization problem is presented and two proposed frameworks are then described. A first SBO algorithm JPAD Optimizer is introduced in Section~\ref{sec:jpad} and a second BO framework SEGOMOE is given in Section~\ref{sec:segomoe} where the Continuous Relaxation (CR) as well as the use of the Partial Least Squares (PLS) technique are detailed. Some details concerning the remote access to the frameworks are also provided.
Then Section~\ref{sec:AC} presents three MDO application cases where multiple physics and domains are involved. For each of the application cases,  multidisciplinary analysis, optimization problem as well as the obtained results are detailed.  
Conclusions and perspectives are finally drawn in Section~\ref{sec:conclu}.

\section{Multi-objective Optimization algorithms}\label{sec:multiobj}
To begin with, we start formulating the multi-objective paradigm over which the paper will be based. Then, we introduce our two frameworks used to solve such multi-objective optimization problems in an industrial context.

\subsection{Multi-objective formalism}
\begin{figure}[!htb]
    \centering
    \includegraphics[scale=0.8]{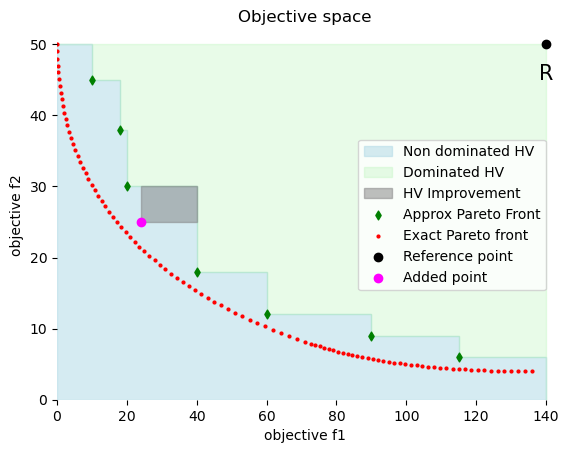}
    \caption{Hypervolume Improvement: the hypervolume indicator of the non-dominated set (green points)
corresponds to the area dominated by it, up to $R$ (reference point in black).
The gray rectangle is the hypervolume improvement brought by the new added
point in magenta.}
    \label{fig:HVI}
\end{figure}

Consider the multi-objective constrained optimization problem
\begin{equation}
\displaystyle
\left\lbrace
\begin{array}{l}
 \displaystyle\min_{  \x \in  \Omega \times S \times \mathbb{F}^l  } \quad \f(\x):= [f_1(\x), f_2(\x), \cdots, f_n(\x)]\\
\mbox{s.t.}  \quad g_{1}(\x) \leq 0\\
\quad \quad \quad  \ \  \vdots\\
\quad  \quad  g_{m}(\x) \leq 0,\\
\end{array}\right.
\label{eq:opt_problem}
\end{equation}
where $ \Omega \subset \mathbb{R}^d$ represents the bounded continuous design set for the $d$ continuous variables, $S \subset \mathbb{Z}^\ell$ represents the bounded integer set where $L_1, \ldots , L_\ell$ are the numbers of levels of the $\ell$ quantitative integer variables on which we can define an order relation, and $ \mathbb{F}^l := \{1, \ldots, L_1\} \times \{1, \ldots, L_2\} \times  \ldots \times \{1, \ldots, L_l\}$ is the design space for $l$ categorical qualitative variables with their respective  $L_1, \ldots, L_l$ levels~\cite{Mixed_Paul}.
The objectives functions are referred as $f_i:\mathbb{R}^d \to \mathbb{R} \quad \forall i=1,\ldots, n$ where $n$ is the number objectives.
The constraints are named $g_j:\mathbb{R}^d \to \mathbb{R} \quad \forall j=1,\ldots, m$ where $m$ is the number of constraints (inequality or equality).
Due to potentially conflicting objectives, the solution of the optimization is not unique but a range of solutions is proposed.
The trade-off between these solutions is characterized by the notion of dominance:
a solution $\x$ is said to dominate another solution $\x'$ and denoted by $\x \preceq \x'$ if
$$f_i(\x) \le f_i(\x') \quad \forall i=1,\ldots, n.$$
The set of solutions representing the optimal trade-off is referred to the Pareto set (PS) and the corresponding image of PS in the objective space is known as the Pareto front~(PF)
$$\mbox{PF}:=\{\f(\x)| \x \in \text{PS} \}.$$
Figure~\ref{fig:HVI} illustrates a Pareto front (red points) relative to two objectives. An approximation to this PF is given by the scattered green dots and the associated dominated hypervolume given by the green area.
Both the SBO and BO frameworks to solve problem~\eqref{eq:opt_problem} are described in the following.

\subsection{Surrogate-Based Optimization with JPAD}\label{sec:jpad}

In this work, one of the optimization tools used is the JPAD Optimizer, which is based on the  Multi-Objective Evolutionary Algorithms (MOEA) framework. It should be noted that the JPAD Optimizer is an independent implementation within the JPAD library, as described in~\cite{nicolosi2016development, trifari2017java}. The MOEA framework is a Java library that serves as an open-source platform for the development and experimentation of various optimization techniques. The latter offers a range of pre-implemented algorithms that can be readily employed, such as genetic algorithms and particle swarm optimization, among others. For this particular study, the $\epsilon$-NSGAII algorithm has been used. The $\epsilon$-NSGAII algorithm extends the NSGA-II algorithm by incorporating an $\epsilon$-dominance archive and randomized restarts, thereby enhancing the search process and enabling the discovery of a diverse set of Pareto optimal solutions. A comprehensive explanation of this algorithm can be found in~\cite{kollat2005comparison} with in-depth details and comparisons. The JPAD Optimizer requires a comprehensive factorial database to conduct the analysis effectively. This database enables the solver to construct a specific Response Surface Model (RSM) through spline interpolation. The RSM serves as a surrogate model, capturing the relationship between the input variables and the response variable. 
By utilizing the RSM as a surrogate model, the optimization algorithm can efficiently explore the design space and identify optimal solutions. However, it is important to note that direct inclusion of categorical variables in the optimization is not feasible. To address this limitation, one potential solution is to perform the optimization iteratively, considering each categorical variable separately for each run. Still, using the JPAD Optimizer in conjunction with the MOEA framework provides a robust and versatile approach to the optimization process. 
The selection of the $\epsilon$-NSGAII algorithm within the MOEA framework allowed for efficient and effective exploration of the design space, leading to the identification of a diverse range of Pareto optimal solutions. The chosen references provide additional information on the JPAD library, the MOEA framework, and the $\epsilon$-NSGAII algorithm, serving as valuable resources for readers interested in a more detailed understanding of the employed methodologies.

\subsection{Bayesian Optimization with SEGOMOE}\label{sec:segomoe}

The second proposed optimization process is based on a sequential enrichment approach, typically the Efficient Global Optimization algorithm~\cite{jones1998efficient} or Super EGO~\cite{Sasena02flexibility}, an evolution of EGO to handle constraints. 
Bayesian Optimization (BO) is based on Gaussian processes (GP)~\cite{rasmussen2006gaussian} (also denoted by Kriging~\cite{krige1951statistical}) and here the idea is to use some adaptive mixture of Kriging based models to tackle high dimension problems. The mixture of experts (MOE)~\cite{dimitri:SM02011, Liem2015} is known to approximate complex functions with heterogeneous behaviour by combining local surrogate models in a global one. In order to consider high-dimensional functions to approximate objective functions and constraints, we used adapted local Kriging-based  models~\cite{Bouhlel2016_SMO, Bouhlel2016_MPE}. Some recent developments have been made to consider highly non linear constraints~\cite{BRAC_AIAA:20}, mixed integer variables~\cite{saves2022multidisciplinary} and multi-objective applications~\cite{savesMOO_AIAA:22}. 
The general framework called Super Efficient Global Optimization coupled with Mixture Of Experts (SEGOMOE) has been proposed by ONERA and ISAE-SUPAERO. The performance of SEGOMOE is validated and proven on different analytical and industrial test cases~\cite{SEGOMOE_AST:18, AGILE_PIAS_MDO:2020, bussemaker2021effectiveness}.

Here we consider multiple objectives and constraints with mixed integer variables. Some continuous, integer and categorical variables are involved within the different application cases, so the GP surrogate models have to be adapted to deal with.  We proposed to combine continuous relaxation to replace categorical variables by continuous ones and Partial Least Squares techniques to handle the increase of dimensionality. The continuous relaxation introduced by~\cite{GMHL} uses a one-hot encoding strategy~\cite{one-hot} and transforms integer and categorical inputs into continuous ones, so some classic continuous correlation kernels can still be used to build the Kriging models. 
The design  space  $ \Omega \times S \times \mathbb{F}^l $ is relaxed to a continuous space $ \Omega'$  constructed on the following way~\cite{saves2022multidisciplinary}:
\begin{itemize}
\item $\forall i \in \{1, \ldots,\ell\}$, the integer variable $z_i$ is relaxed within its bounds and treated as continuous. 
\item   $\forall j \in \{1, \ldots,l\}$, we use a relaxed one-hot encoding~\cite{one-hot} for the categorical variable $c_j$ (and its $L_j$ 
 associated levels) and add $L_j$ new continuous dimensions into $ \Omega'$. 
\end{itemize}
Therefore, we get, after relaxation, a new design space $\Omega'\subseteq \mathbb{R}^{d'}$ where $d'= d+\ell+ \sum_{j=1}^l L_j >d+\ell+l$. 
As the number of variables may increase drastically (from $d$ to $d'$), one new added variable associated to each categorical level, PLS technique is combined with Kriging, denoted as KPLS and KPLS-K models~\cite{Bouhlel2016_SMO, Bouhlel2016_MPE}, to reduce the number of hyperparameters. During experiments, the chosen number of principal components within KPLS models does not exceed 4 or 5.  The code implementation of the proposed GP has been released in the toolbox SMT v1.1\footnote{\url{https://smt.readthedocs.io/en/latest/}}~\cite{SMT2019}. 

Once the initial Design of Experiments is given in the relaxed space $\Omega'$, Kriging based surrogate models are built for the objective and constraint functions.  Each costly function $f_i(\x)$ is approximated by a GP characterized by its mean $\hyi: \mathbb{R}^{d'} \to \mathbb{R}$ and its standard deviation ${s_i}(\x): \mathbb{R}^{d'} \to \mathbb{R}$
    $$\hat{f}_i(\x)\sim \mathcal{N}(\hyi,\hsi) \quad i=1,\ldots, n$$
For the multi-objective, we assume that the $n$ components of $\f$ are independant to define $\hat{\f}$ as the surrogate model  associated to each component 
$$\hat{\f}(x)\sim \mathcal{N}(\hat{\y}(\x),\Sigma(\x)) $$
where $\hat{\y}(\x): \mathbb{R}^{d'} \to \mathbb{R}^n$ is the GP prediction vector given by $[\hat{y}_1(\x),\ldots\hat{y}_n(\x)]$ and $\Sigma(\x)$ is a diagonal matrix whose diagonal vector is given by $\hsi, \forall i=1,\ldots, n$.
So the initial problem defined by~\eqnref{eq:opt_problem} is replaced by an infill problem defined as follows
\begin{equation}\label{eq:opt_acquisition_problem}
\left\lbrace
\begin{array}{l}
 \displaystyle \max_{{x}\in \mathbb{R}^{d'}} \quad \alpha^{\mbox{reg}}_{\f}(\x) \\
\mbox{s.t.}  \quad \hat{g}_{1}(\x) \leq 0\\
\quad \quad \quad  \ \  \vdots\\
\quad  \quad  \hat{g}_{m}(\x) \leq 0,\\
\end{array}\right.
\end{equation}
where $\hat{g_i}(\x)$ corresponds to the mean prediction of the GP constraint models and the regularized acquisition function~\cite{savesMOO_AIAA:22} is defined by
\begin{eqnarray} \label{eq:alpha:reg}
\alpha^{\mbox{reg}}_{\f}(\x) &:=& \gamma\;\alpha_{\f}(\x) -\psi(\mu_{\f}(\x))
\end{eqnarray}
where $\alpha_{\f}(\x)$ is a standard scalar acquisition (EHVI, PI, MPI, \dots) depending on $\hat{\y}(\x)$ and $\Sigma(\x)$,
and $\gamma$ is a constant parameter. The function $\psi: \mathbb{R}^{d'} \to \mathbb{R}$ is a scalarization operator. Different choices exist for the function $\psi$ and in~\cite{savesMOO_AIAA:22} two options were investigated. Namely, for a given $\hat{\y}(\x)\in \mathbb{R}^{d'}$, we consider
\begin{eqnarray*}
(\mbox{reg}=\max)&: & \psi(\hat{\y}(\x)) :=  \max_{i \leq d'}\hyi  \label{eq:max}\\
(\mbox{reg}=\mbox{sum}) & :& \psi(\hat{\y}(\x)):= \sum_{i=1}^{d'} \hyi \label{eq:sum}
\end{eqnarray*}
The performance of this regularized acquisition function~\eqref{eq:alpha:reg} has been tested on analytical examples involving continuous variables and apply with success to an aircraft design test case in~\cite{savesMOO_AIAA:22}.

Concerning the different BO criteria relative to the hypervolume Improvement, an illustration is proposed in~\figref{fig:HVI}. The Expected Hyper-Volume Improvement (EHVI~\cite{zitzler2003performance,emmerich2006single}), the Probability of Improvement (PI~\cite{jones2001taxonomy}), and the Minimum of Probability of Improvement (MPI~\cite{rahat2017alternative}) are some multi-objective extensions of the well known Expected Improvement (EI~\cite{jones1998efficient}).
As seen in~\figref{fig:HVI}, with respect to the predictions in the Gaussian random field, the idea is to measure how much hypervolume improvement (grey area) could be achieved by evaluating the new point (magenta point), considering
the uncertainty of the prediction. These criteria differ depending how they favor well-spread solutions: for instance for EHVI the
hypervolume increase is small when adding a new value close to an already
observed one in the objective space.
To solve problem~\eqnref{eq:opt_acquisition_problem}, any optimization algorithms capable of considering non linear constraints based
either on derivative free optimizer such as COBYLA  (Constrained Optimization BY Linear Approximation~\cite{powell1994direct}) or
based on gradient method such as SLSQP (Sequential Least Squares Programming~\cite{kraft1988software}) or SNOPT (Sparse Nonlinear
OPTimizer~\cite{gill2005snopt}) can be used together with a multistart strategy.
This adaptive process is repeated until the total budget is reached. The feasible points of the final database represent the known Pareto optimal points. Nevertheless, as the set of known points has been enriched sequentially to increase the hypervolume, the final database can be used to build GP models for objectives and constraints as a post-processing step. These final GP models can be coupled to an evolutionary algorithm to deal with the multi-objective constrained problem and retrieve the approximated PF. The well known NSGA-II algorithm (Non-dominated Sorting Genetic Algorithm II~\cite{nsga2}) of the pymoo~\cite{pymoo} toolbox\footnote{\url{https://pymoo.org/}} is used for that purpose  and ultimately provides the predicted PF based on GP with almost no additional computational cost.

In the end, the proposed strategy provides two outputs: the PF database and the predicted PF. Comparing the proximity between the two PF is a good criterion to know if some additional enrichment points are needed or if the accuracy is sufficient.  
Algorithm~\ref{algo:SEGOMOE_Multiobj} details the SEGOMOE optimization procedure. 

\smallskip
\begin{algorithm}[H]
\SetAlgoLined
{\textbf{Inputs:}}  Initial DoE $\mathscr{D}_0$ and set $t=0$;   

\While{the stopping criterion is not satisfied}{
\vspace{.2cm}
\begin{adjustwidth}{0pt}{40pt}
\begin{enumerate}
    \item  Relax continuously integer and categorical input variables to a real bounded space $\Omega'$ of dimension $d'= d+\ell+ \sum_{j=1}^l   L_j$. Namely, we continuously relax the mixed categorical DoE to a continuous DoE $\mathscr{D}_t$ using the relaxation procedure;
    \item Build the GP model for each objective function $f_i(\x)$ and each constraint $g_j(\x)$ related to the continuous DoE with PLS to reduce the number of the hyperparameters and compute an estimation of the search space $\Omega_f$;
    \item Build the acquisition function $\alpha^{\mbox{reg}}_{\f}(\x)$;
    \item Maximize the acquisition function within the feasible domain  $\Omega_f$: $$\x_{t}:=  \underset{\x \in \Omega_f}{\arg \max}  \  \alpha^{\mbox{reg}}_{\f}(\x)$$
    \item Add $\x_{t}, \f(\x_{t}) $ and $ g_1(\x_{t}), \ldots g_m(\x_{t})$ to the DoE  $\mathscr{D}_{t+1}$. Increment $t$;
\end{enumerate}
\end{adjustwidth}
\smallskip
} 
{\textbf{Post-process}}: Use the final database to build GP models (for $f_i(\x)$ and $g_j(\x)$) and apply NSGA-II algorithm to construct the PF ;\\
{\textbf{Outputs:}}  The PF database and the predicted PF; 
\caption{SEGOMOE for constrained multi-objective and mixed-integer problems.}
 \label{algo:SEGOMOE_Multiobj}
\end{algorithm}

\smallskip

\subsection{Optimizers access}\label{sec:access}
In AGILE 4.0 project, a Process Integration and Design Optimization (PIDO) component, also known as RCE~\cite{boden2021rce}, is required to execute disciplinary workflows. This executable workflow integrates the different disciplinary tools, which are treated as black-boxes: only inputs and outputs are relevant for the component and the codes are not exposed. The disciplinary tools are installed locally or in servers. Different types of MDO problems can be executed in the component, as basic multidisciplinary analysis, DOEs or full optimizations. 
Another key aspect of the collaborative workflow derives from the collaborative design process. Since the different disciplinary tools can be hosted in different locations, executable workflows are typically distributed simulation workflows across multiple networks in multiple companies. In AGILE 4.0, Brics tool~\cite{baalbergen2017methods,baalbergen2022advancing} is used to provide the mechanism for interconnecting PIDO environments and hence multi-partner and distributed collaborative simulation workflows. By providing data encryption services, Brics supports protection of the data to be exchanged to and from the central data server in which the inputs/outputs (I/O) file is transferred between the disciplinary tools. In the project, I/O file relies on Common Parametric Aircraft Configuration Schema (CPACS)~\cite{alder2020recent}. CPACS is an open-source, XML-based common language for the exchange of product data. It allows storage of parametric definitions of aircraft geometries as well as analysis results of the individual design disciplines. Therefore, any legacy engineering tool in AGILE 4.0 context is equipped with a wrapper that enables the tool to act on product data and produce analysis results in CPACS format.

As all the application cases have to address multi-objective optimization problems with mixed variables, the optimization algorithms currently available through RCE were not entirely suitable (direct application of genetic algorithms would result in an excessively large number of evaluations). Only a few partners, such as ONERA, NLR, or UNINA, were able to provide gradient-free algorithms capable of solving these types of optimization problems.  
Therefore three different technical approaches have been applied to connect optimizers to the workflows:
\begin{itemize}
\item Local optimizer: the first approach consisted in using an adequate optimization algorithm located on the integrator’s site. This means that the integrator of the workflow needs to have also an expertise in optimization. 
In AGILE 4.0, only UNINA could apply this approach with JPAD algorithm as illustrated in Section~\ref{sec:AC6}.
\item Remote optimizer through Brics call: the second approach consisted in considering an MDA workflow as a whole that can be called remotely like any design competence from an RCE specific workflow located in the optimizer owner’s premises. This approach was successfully applied in multiple application cases and was retained for the application case presented in Section~\ref{sec:AC2}.
\item Remote access through WhatsOpt: the third approach applied the optimization process not on the design workflow but directly on the architectural choices that is connected to the RCE workflow. In that case, the optimizer is made accessible “as a service” through WhatsOpt~\cite{lafage2019whatsopt}, a web application supporting MDO collaborative activities. This approach was successfully demonstrated in the application case illustrated in Section~\ref{sec:AC7}.
\end{itemize}

%
\section{Applications}\label{sec:AC}
\subsection{Airframe upgrade design}\label{sec:AC6}
\begin{figure}[!htb]
  \centering
  \includegraphics[width=\textwidth]{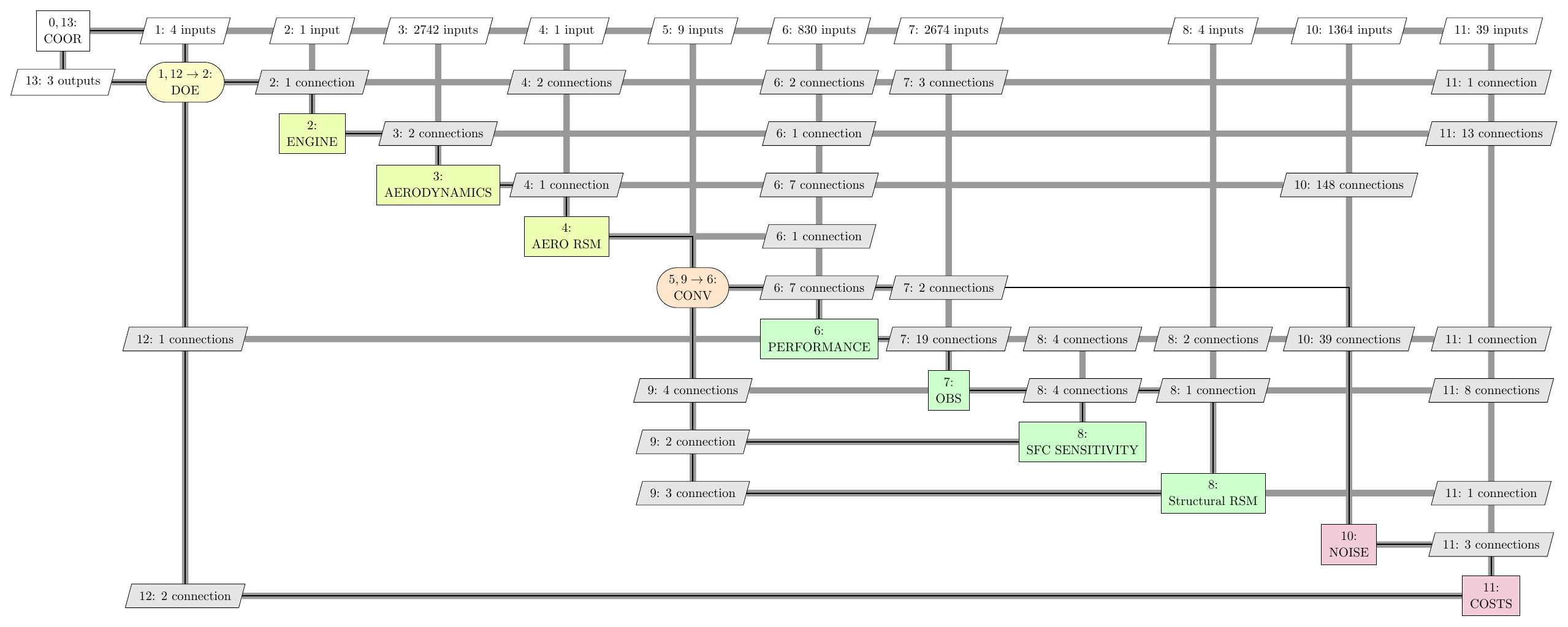}
  \caption{XDSM DOE MDA for Airframe upgrade design.}
  \label{fig:MDA_AC6}
\end{figure}

This first application case is part of the Upgrade driven stream and aims at accounting retrofitting options in the aircraft MDO workflow. The AGILE 4.0 MDO framework is used to design various retrofitting concepts, including engine replacement, OBS electrification, and winglet installation. Each solution has different impacts on aircraft performance, such as fuel consumption, direct operating cost (DOC), air and noise emissions, as well as different retrofitting costs. The objective is to be able to select the optimal retrofitting strategy by trading DOC and emissions against retrofitting costs. \\
For this specific application case, a classical Multidisciplinary Feasible (MDF) approach is being utilized. This type of architecture involves an optimizer driving an MDA workflow, where a complete design is executed in each iteration. The MDA components comprising the workflow can be observed in Figure~\ref{fig:MDA_AC6}. The workflow begins with an engine module, which provides key engine characteristics based on the given engine ByPass Ratio (BPR) input. This module relies on GASTurb~\footnote{https://www.gasturb.de/ GasTurb} computation. The aerodynamics branch calculates values for both low-speed and high-speed conditions. To account for high-fidelity results in high-speed conditions obtained through CFD analysis, a RSM has been developed. The tool also considers engine geometry and position as inputs. The OBS discipline, named ASTRID~\cite{chiesa2012astrid}, is responsible for sizing the overall on-board system (OBS). It determines the masses and the hydraulic, pneumatic, and electric power required by each system during different phases of the mission profile. It also calculates the power-off-takes (secondary power computation) to consider their impact on engine fuel flow. In this study, four different OBS architecture electrification levels are considered, ranging from CONV (conventional) to AEA (all electric aircraft) with two intermediate architectures, namely MEA1 and MAE2 (more electric aircraft 1 and 2). A more detailed description of these architectures can be found in~\cite{della2018advanced}. The performance tool uses a simulation-based approach to compute ground and in-flight performance. It calculates overall mission profile, performance, fuel consumption, flight time, and gaseous emissions. The structural competence relies on a surrogate model based on a high-fidelity structural analysis DOE. This tool optimizes the wing mass by adjusting lamination parameters and thicknesses while ensuring aeroelastic stability, angle-of-attack, strength, and buckling constraints are met. The last three tools described are executed in a converger loop, with iterations performed until convergence is achieved on the maximum take-off weight. The noise competence calculates noise emissions at the certification points in accordance with FAR36 and ICAO Annex 16 regulations. It also provides the noise margin with respect to the certification limits. The cost competence evaluates both recurring and non-recurring costs, including aircraft price and direct operating costs. It also estimates the development, operation, and equipment costs associated with a retrofitting activity. Additionally, it computes cost savings (a component of direct operating costs) resulting from reductions in fuel consumption, maintenance costs, and emission taxes. The cost methodology is explained in~\cite{della2022retrofitting}. Further details about the MDA workflow presented here can be found in~\cite{mandorino2022multidisciplinary}.

\begin{figure}[!htb]
  \centering
  \includegraphics[width=\textwidth]{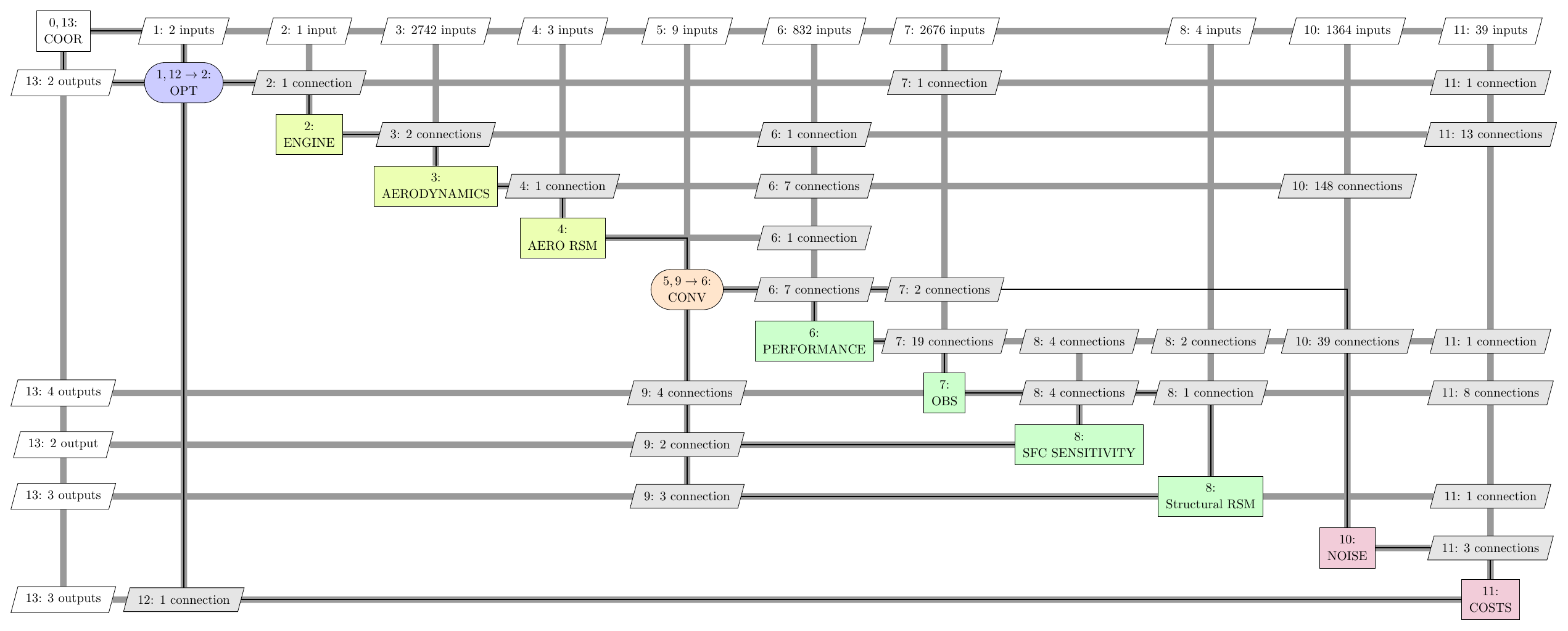}
  \caption{XDSM optimization for airframe upgrade design.}
  \label{fig:OPT_AC6}
\end{figure}

This application case includes the integration of costly evaluations required to compute the wing structural weight and the high-fidelity aerodynamic effect of the relative nacelle/wing location. To mitigate the cost and computational time, for each design competence, a RSM was built in an off-line process with SMT~\cite{SMT2ArXiv}. A database computed by PROTEUS~\cite{werter2016novel}, a tool developed by TUD, is used to calculate the wing's structural mass. The RSM utilizes a database of 56 points (to achieve more than 10 points per dimension) obtained through a Latin Hypercube Sampling (LHS) approach. Additionally, an RSM has been developed based on CFD analysis provided by CFSE to consider the aerodynamic effects of both the pylon and the installed engine on the drag coefficient. The RSM for this purpose utilizes a database of 90 points obtained through an LHS approach. For this RSM, the initial database was smaller (around 10 points per dimension) but had to be enriched to increase the accuracy of drag prediction.  These RSMs have been incorporated as design competences within this workflow.\\
Regarding the choice of optimization solver, two different approaches were tested during the frame of the project. The first optimization strategy relies on the JPAD optimizer provided by UNINA and described in Section~\ref{sec:jpad}. In this approach, a DOE (full factorial composed of 108 points) is built using the workflow illustrated in Figure~\ref{fig:MDA_AC6}. In a second step, using JPAD Optimizer, four different optimizations are performed, each corresponding to one of the categorical variables (OBS choice) considered in the optimization problem. For each of this optimization and each quantity of interest, a specific RSM using spline interpolation is built based on the obtained database using the JPAD optimizer. Subsequently, the optimization algorithm is applied, using the RSM. The generation of the DOE dataset occurs online, while the offline optimization process takes place once all the data are available for processing. The second optimization approach takes advantage of SEGOMOE, from ONERA (see Section~\ref{sec:segomoe}) that can handle directly categorical variables. With this multi-objective BO solver, a remote call to the MDA workflow from the ONERA location was necessary at each evaluation as explained in Section~\ref{sec:access}. The optimization workflow executed using SEGOMOE is illustrated in Figure~\ref{fig:OPT_AC6} starting with an initial DOE of 13 points (selected among the  108 full factorial points) and running 68 iterations.

\begin{figure}[!htb]
   \begin{subfigure}[b]{.5\linewidth}
      \centering
      \includegraphics[width=\linewidth]{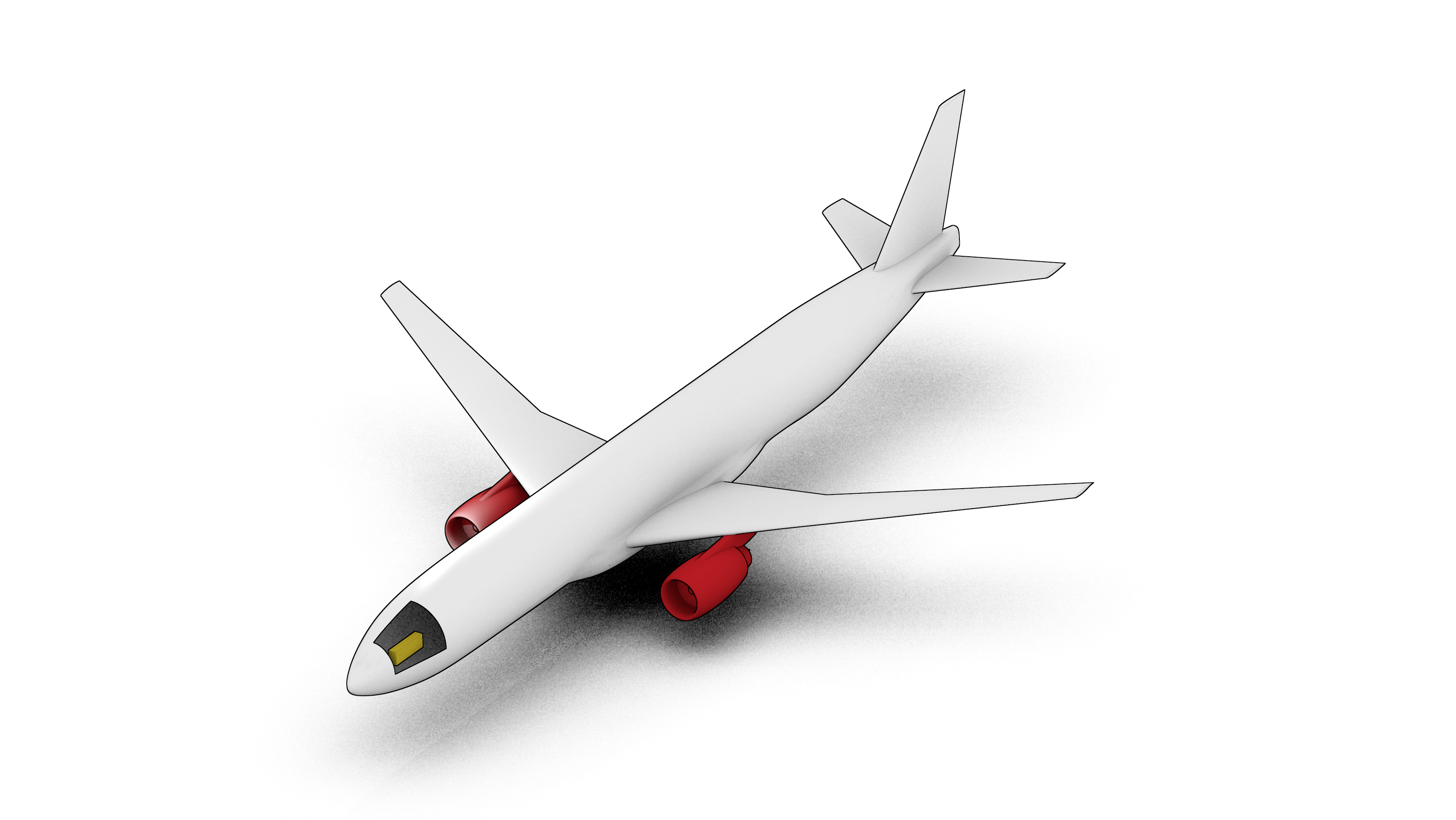}
      \caption{Engines and OBS are highlighted.}
      \label{fig:myfig1}
      \end{subfigure}
      \begin{subfigure}[b]{.5\linewidth}
      \centering 
      \scalebox{-1}[1]{\includegraphics[width=5cm]{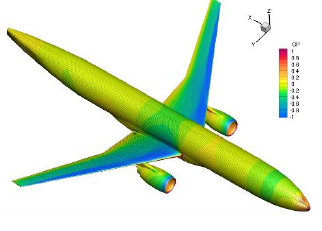}}
      \caption{CFD simulation of nacelle location effects.}
      \label{fig:myfig2}
   \end{subfigure}
   \caption{Reference regional jet aircraft.}
   \label{fig:Referece_AC_AC6}
\end{figure}

The optimization problem addressed by both of the methodologies presented is defined in Table~\ref{tab:pb_ac6}. Four main objective functions have been established. The first objective is the maximum take-off weight (MTOW) of the aircraft, which is computed through the iterative convergence process described earlier. The second objective is the cumulative emission index (CEI), which quantifies the level of emissions in terms of air and noise. This index is calculated as a weighted sum of three emissions, expressed in non-dimensional units. The third objective is the difference between costs and savings. This objective captures the disparity between two economic factors. The first factor encompasses the capital costs incurred by the manufacturer to retrofit the aircraft, including the expenses associated with development, operations, and equipment required for the retrofitting activity. The second factor concerns the operational savings, which include the reductions in fuel consumption, maintenance costs, and emission taxes achieved compared to the baseline aircraft during a typical mission. The fourth objective is the maximum specific air range (SAR) achievable in cruise conditions and weight, representing the aircraft's flight efficiency and fuel consumption. Four design variables are considered. One variable is categorical, representing four different OBS architectures that range from conventional to all-electric configurations. The other three variables are continuous, including the engine BPR and the engine $X$ and $Z$ locations with respect to the wing attachment points. Figure~\ref{fig:Referece_AC_AC6} represents the system under analysis, illustrating the components that can be retrofitted and are affected by the design variables considered in the problem. Lastly, four constraints have been selected. The retrofit solution must maintain at least the same maximum take-off weight and the same take-off and landing distances as the baseline configuration. Additionally, a reduction of 6 $dB$ in cumulative noise emissions must be achieved compared to the original aircraft, leading to a maximum cumulative noise equal to 263 $dB$.


\begin{table}[!htb]
   \caption{Definition of  the airframe upgrade design problem.}
   \small
   \resizebox{\linewidth}{!}{
      \begin{tabular}{lclrr}
 & Function/variable & Nature & Quantity & Range\\
\hline
\hline
Minimize & Maximum Take-Off Weight & cont & 1 &\\
Minimize &  Cumulative Emission Index & cont & 1 &\\
Minimize &  Cost - savings & cont & 1 &\\
Maximize &  Max Cruise Specific Air Range & cont & 1 &\\
 & \multicolumn{2}{l}{\bf Total objectives} & {\bf 4} & \\
\hline
with respect to & OBS architecture & cat  &  4 levels & [CONV, MEA1, MEA2, AEA]\\
 & \multicolumn{2}{l}{ Total categorical variables} & { 1} & \\ \cline{2-4}
 & Engine Bypass Ratio & cont & 1 & $\left[9,15 \right]$ ($-$)\\
 & Engine $X$ & cont & 1 & $\left[ -0.98, -0.80  \right]$ ($m$)\\
 & Engine $Z$ & cont & 1 & $\left[-0.39,-0.21\right]$  ($m$)\\
 & \multicolumn{2}{l}{ Total continuous variables} & { 3} & \\ \cline{2-4}
 & \multicolumn{2}{l}{\bf Total relaxed variables} & {\bf 7} & \\
\hline
subject to & 
\multicolumn{2}{c}{ Maximum Take-Off Weight $\le 39058.5~kg$ } & 1 & \\
&\multicolumn{2}{c}{ Take-Off Field Length $ \le 1500~m$ }  & 1 & \\
&\multicolumn{2}{c}{ Landing Field Length $\le 1400~m$ }  & 1 & \\
&\multicolumn{2}{c}{ Cumulative Noise $ \le 263~dB$ } & 1 & \\
%
 & \multicolumn{2}{l}{\bf Total constraints } & {\bf 4} & \\
\hline
\end{tabular}
}
   \label{tab:pb_ac6}
\end{table}

\begin{figure}[!htb]
   \begin{subfigure}[b]{.5\linewidth}
      \centering
      \includegraphics[width=\textwidth]{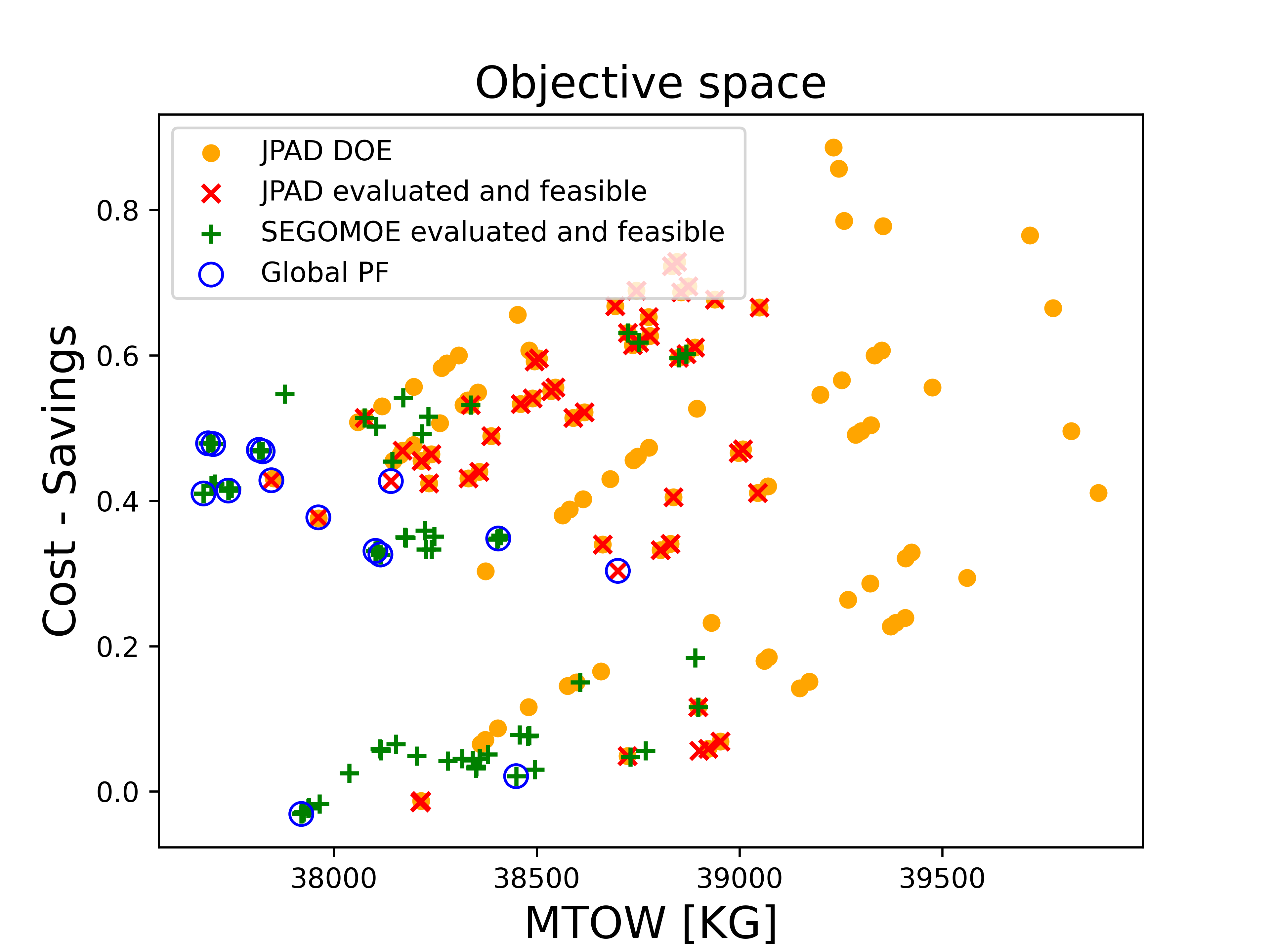}
      \caption{Cost - Savings vs MTOW}
      \label{fig:myfig1}
      \end{subfigure}
      \begin{subfigure}[b]{.5\linewidth}
      \centering 
     {\includegraphics[width=\textwidth]{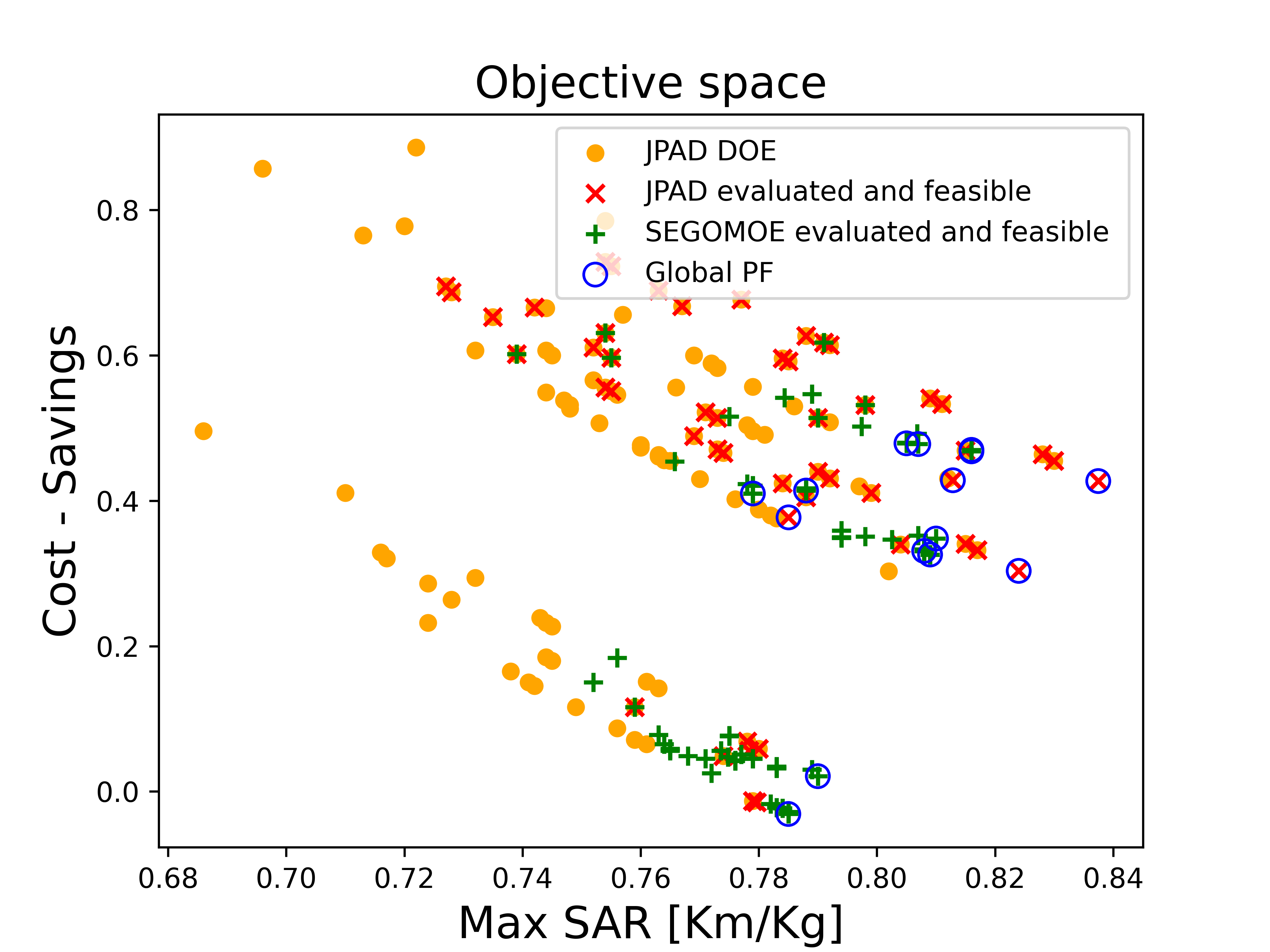}}
      \caption{Cost - Savings vs SAR}
      \label{fig:myfig2}
   \end{subfigure}
   \begin{subfigure}[b]{\linewidth}
      \centering 
      {\includegraphics[width=0.5\textwidth]{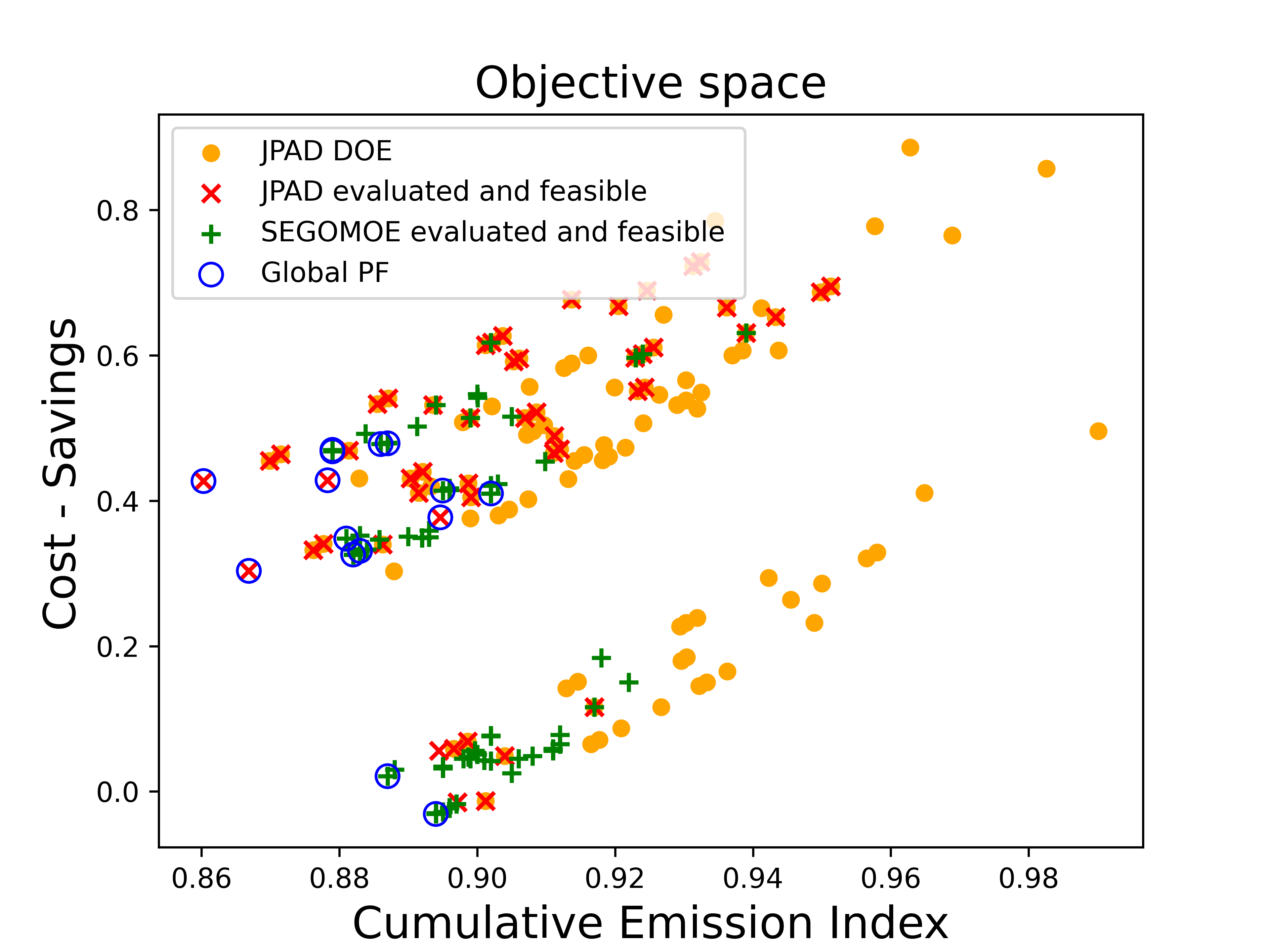}}
      \caption{Cost - Savings vs CEI}
      \label{fig:myfig3}
   \end{subfigure}
   \caption{Different PF using four objectives:  108 DOE points for JPAD (orange dot), 15 points (blue circle) on the final PF combining JPAD (red cross) and SEGOMOE (green cross) databases.
   }
   \label{fig:AC6_PF}
\end{figure}

The following are the results of the optimization process explained earlier. Figure~\ref{fig:AC6_PF} displays the DOE points and the PF obtained using both approaches. The workflow depicted in Figure~\ref{fig:MDA_AC6} generated a total of 108 points in the DOE, represented by orange dots in Figure~\ref{fig:AC6_PF}. On the same figure are highlighted, as red crosses, the  points satisfying the optimization constraints for both DOE (45 feasible points out of 108 full factorial DOE points) and JPAD optimization. 

Regarding the results obtained with SEGOMOE (the workflow shown in Figure~\ref{fig:OPT_AC6}), green crosses represent all the evaluated points fulfilling the optimization constraints (among the 81 points computed with this approach).
The global PF, indicated by the 15 blue circles, was derived from combining both databases obtained by JPAD optimizer and SEGOMOE. This front is composed by 4 points from JPAD optimizer and 11 points from SEGOMOE optimizer. Among the points on the PF, the best solution depends on the designer's preferences. For instance, the point with the minimum difference between costs and savings (the lower one) is always an optimum choice in terms of economic profit. However, it may not be the optimal solution concerning other variables such as CEI and SAR. In fact, several solutions that offer significant benefits in terms of emissions, SAR, and MTOW may be disregarded in the pursuit of achieving lower costs. Similarly, there are points that offer maximum benefits in terms of other variables but may not be desirable due to their higher costs. When examining opposite solutions on the PF, the selection of engine BPR and OBS architecture can vary depending on the performance objective to be maximized: for example, a BPR of 9.0 with a conventional OBS architecture (CONV) or a BPR of 15.0 with an advanced electric aircraft (AEA) OBS architecture. Opting for lower BPR and a moderate level of electrification (representing state-of-the-art technologies) helps reduce retrofitting costs while achieving moderate performance improvements. Conversely, increasing the level of retrofitting by adopting advanced engine technology and full OBS electrification (beyond state-of-the-art) can significantly enhance overall performance metrics like SAR and CEI, but at the cost of higher retrofitting expenses. For instance, considering a higher BPR and advanced electrification, emissions are reduced (CEI decreases from 1 to 0.86), MTOW slightly decreases (around -3.1\% compared to the baseline), and SAR increases (around +25\% compared to the baseline). As a result, the costs minus savings increase to 0.44 million euros per year, indicating a loss for an airline operating the aircraft under the considered scenarios.

In this application, two approaches were compared, both providing solutions located on the PF. For this case, SBO with JPAD appears to be a bit more costly - in terms of number of function evaluations - than BO with SEGOMOE and ends up with a smaller number of feasible solutions. Nevertheless, this approach is less sensitive to a change in optimization problem (in terms of objectives and constraints) as it relies on an initial database whereas SEGOMOE enrichment process is tailored for a unique optimization.

\subsection{Design of an aircraft family}\label{sec:AC7}
This application case is part of the upgrade-driven stream and aims at designing a family of three business jet aircraft considering commonality options in the MDO workflow~\cite{bussemaker2022ac7}. The commonality options enable sharing one or more aircraft components between the aircraft in the family: by sharing components, design, certification, production, and maintenance costs are reduced. However, operating costs might be increased due to the use of components not designed for the typical flight conditions.

\begin{figure}[!htb]
      \centering
      \includegraphics[width=\linewidth]{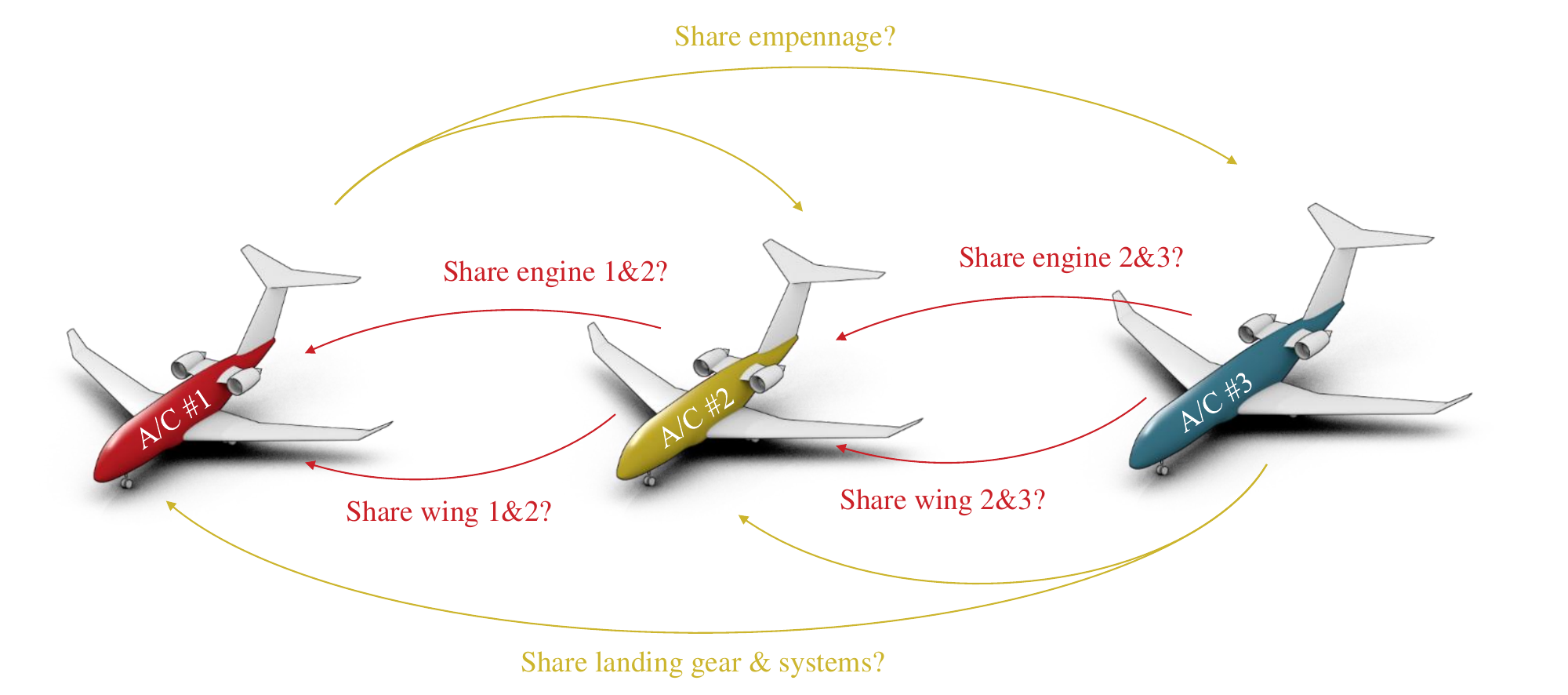}
   \caption{Visualization of the aircraft family including commonality sharing decisions.}
   \label{fig:ac7_sharing}
\end{figure}

Shareable components include wings, empennage (horizontal and vertical tail), engines, on-board systems (OBS), and landing gears, shown in Figure~\ref{fig:ac7_sharing}. Next to the commonality choices there are three design variables per wing: leading edge sweep, rear span location (determining flap size), and thickness-to-chord ratio. These three design variables per wing are inactive if the associated wing is a shared wing. Each aircraft family is optimized for two objectives: direct operating costs (DOC), representing the impact on performance, and Original Equipment Manufacturer (OEM) non-recurring costs, representing the potential benefit of commonality. Table~\ref{tab:pb_ac7} presents the optimization problem in more details.

\begin{table}[!htb]
   \caption{Definition of the aircraft family design problem.}
   \small
      \begin{tabular}{lclrr}
 & Function/variable & Nature & Quantity & Range\\
\hline
\hline
Minimize & Direct Operating Costs & cont & 1 &\\
Minimize &  OEM Non-Recurring Costs& cont & 1 &\\
 & \multicolumn{2}{r}{\bf Total objectives:} & {\bf 2} & \\
\hline
with respect to & engine commonality  1\&2 & cat  &  2 levels & sharing yes/no\\
&engine commonality  2\&3 & cat  &  2 levels & sharing yes/no \\
&wing commonality  1\&2 & cat  &  2 levels & sharing yes/no\\
&wing commonality  2\&3 & cat  &  2 levels & sharing yes/no\\
&landing gear commonality 1\&2 & cat  &  2 levels & sharing yes/no\\
&landing gear commonality 2\&3 & cat  &  2 levels & sharing yes/no\\
& OBS commonality 1\&2 & cat  &  2 levels & sharing yes/no\\
& OBS commonality 2\&3 & cat  &  2 levels & sharing yes/no\\
& empennage commonality  1\&3 & cat  &  2 levels & sharing yes/no\\
& empennage commonality  3\&2 & cat  &  2 levels & sharing yes/no\\
& \multicolumn{2}{r}{ Total categorical variables:} & { 10} & \\ \cline{2-4}
 &  Leading edge sweep* & cont & 3 & $\left[ 30.0, 42.0 \right]$ ($deg$)\\
 & Rear span location* & cont & 3 & $\left[ 0.72, 0.82 \right]$ ($\% chord$)\\
 & Wing thickness/cord ratio*  & cont & 3 & $\left[0.06, 0.11\right]$  ($-$)\\
 & \multicolumn{3}{l}{* only active if the corresponding wing is \textbf{not} shared} \\
 & \multicolumn{2}{r}{Total continuous variables:} & { 9} & \\ \cline{2-4}
 & \multicolumn{2}{r}{\bf Total relaxed variables:} & {\bf 29} & \\
\hline
subject to & 
\multicolumn{2}{c}{ Balanced  Field Length $\le 1524~m$ } & 1 & \\
&\multicolumn{2}{c}{ Landing Field Length $ \le 762~m$ }  & 1 & \\
 & \multicolumn{2}{r}{\bf Total constraints:} & {\bf 2} & \\
\hline
\end{tabular}
   \label{tab:pb_ac7}
\end{table}

\begin{figure}[!htb]
  \centering
  \includegraphics[width=\textwidth]{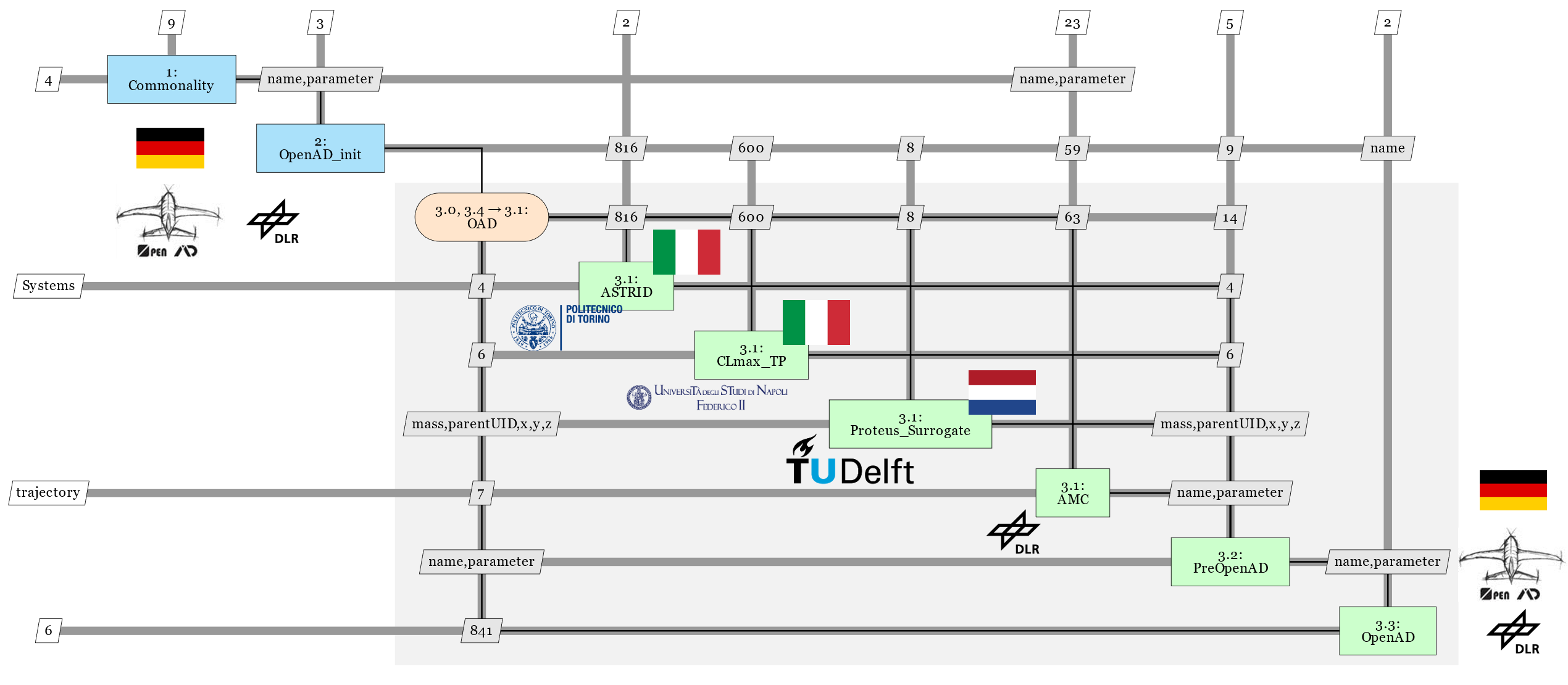}
  \caption{MDA for family aircraft design.}
  \label{fig:MDA_AC7}
\end{figure}

The design space is modeled using the Architecture Design Space Graph (ADSG) method~\cite{bussemaker2020adsg} implemented in the ADORE tool~\cite{bussemaker2022adore}, both developed during the project. MultiLinQ is used to couple the generate architecture instances to the central data schema used in the MDO workflow~\cite{bussemaker2022ac7}. The MDO workflow consists of two levels: an aircraft-level workflow (shown in Figure~\ref{fig:MDA_AC7}) applying commonality decisions and sizing one aircraft at a time, and a family-level workflow converging the three aircraft-level workflows and performing cost calculation on the family-level. For a more detailed overview of the design space model, coupling to the MDO workflow and implementation of the MDO workflow for the family design application case, the reader is referred to~\cite{bussemaker2022ac7,agile4is2023}.

For this application case SEGOMOE is used as optimization algorithm due to the need to handle hierarchical, mixed-discrete design variables, and the need to minimize the number of function evaluations as one family evaluation can take up to 2 hours. Hierarchical design variables were handled using the imputation method, where inactive variables are replaced by the mean value to prevent duplicate design vectors~\cite{bussemaker2021effectiveness}. SEGOMOE was accessed through an ask-tell API implemented in WhatsOpt running on a server at ONERA's premises as described in Section~\ref{sec:access}. For a more detailed description of how SEGOMOE was coupled to ADORE and the MDO workflow, the reader is referred to~\cite{agile4is2023}.

\begin{figure}[!htb]
      \centering
      \includegraphics[width=\linewidth]{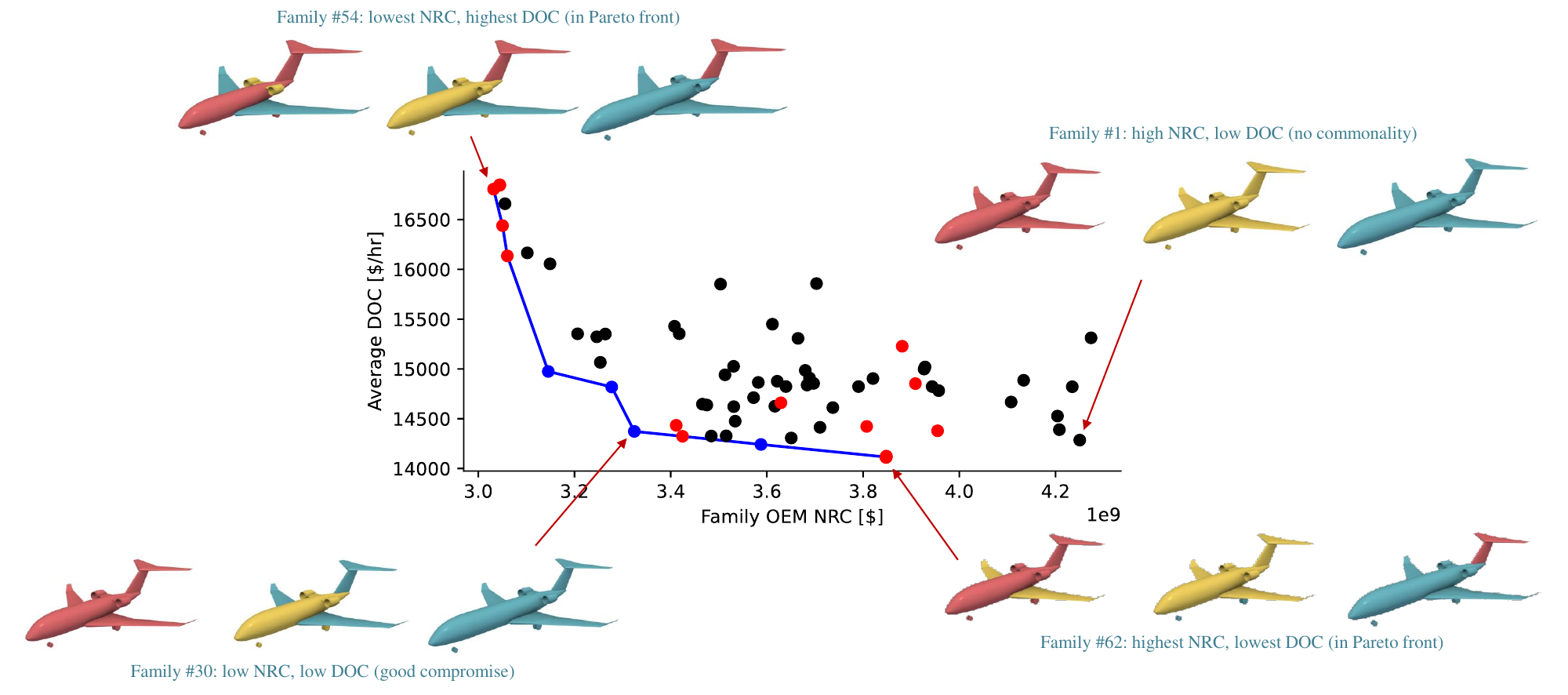}
   \caption{Results of the aircraft family design, showing the Pareto front for minimization of Original Equipment Manufacturer Non-Recurring Cost and Direct Operating Cost and several families. The initial DOE is given by the 50 black dots. Infill points generated by SEGOMOE are shown in red. Colors correspond to the originating aircraft as defined in Figure~\ref{fig:ac7_sharing}.}
   \label{fig:ac7_results}
\end{figure}

First, a 50-point DOE was executed to create the initial database of design points for SEGOMOE and to verify the correct behavior of the MDO workflow. Then, SEGOMOE was used to generate an addition of 18 infill points to explore the design space and extend the Pareto front. Figure~\ref{fig:ac7_results} shows the main Pareto front, with infill points shown in red. Several example families are shown with colors indicating the originating family member (colors are defined in Figure~\ref{fig:ac7_sharing}). As can be seen, when no components are shared (family  $\#1$) the OEM non-recurring costs are high whereas the operating costs are low, because all aircraft components are used at the operating points they are designed for. Introducing more component sharing reduces OEM NRC: family $\#62$ achieves the lowest operating costs at a reduced NRC. Family $\#54$ represents the opposite extreme: the lowest NRC coupled with the highest operating costs, as achieved by a high number of shared components. Family $\#30$ represents a good compromise, at a reduced NRC compared to family $\#62$, with only a moderate increase in DOC.

\subsection{Aircraft design with supply chain, performance  and manufacturing process}\label{sec:AC2}
This application case is part of the Production stream and aims at accounting the supply chain aspects in the aircraft MDO workflow, in addition with design and manufacturing aspects. 
The application  focuses on the design of a horizontal tail plane (HTP) made by different materials and produced by different supply chains, combination of enterprises.
The different solutions are characterized by different impact on aircraft performance (e.g. mass, aerodynamic efficiency), different manufacturing processes and different combinations of supply chains.

\begin{figure}[!htb]
  \centering
  \includegraphics[width=\textwidth]{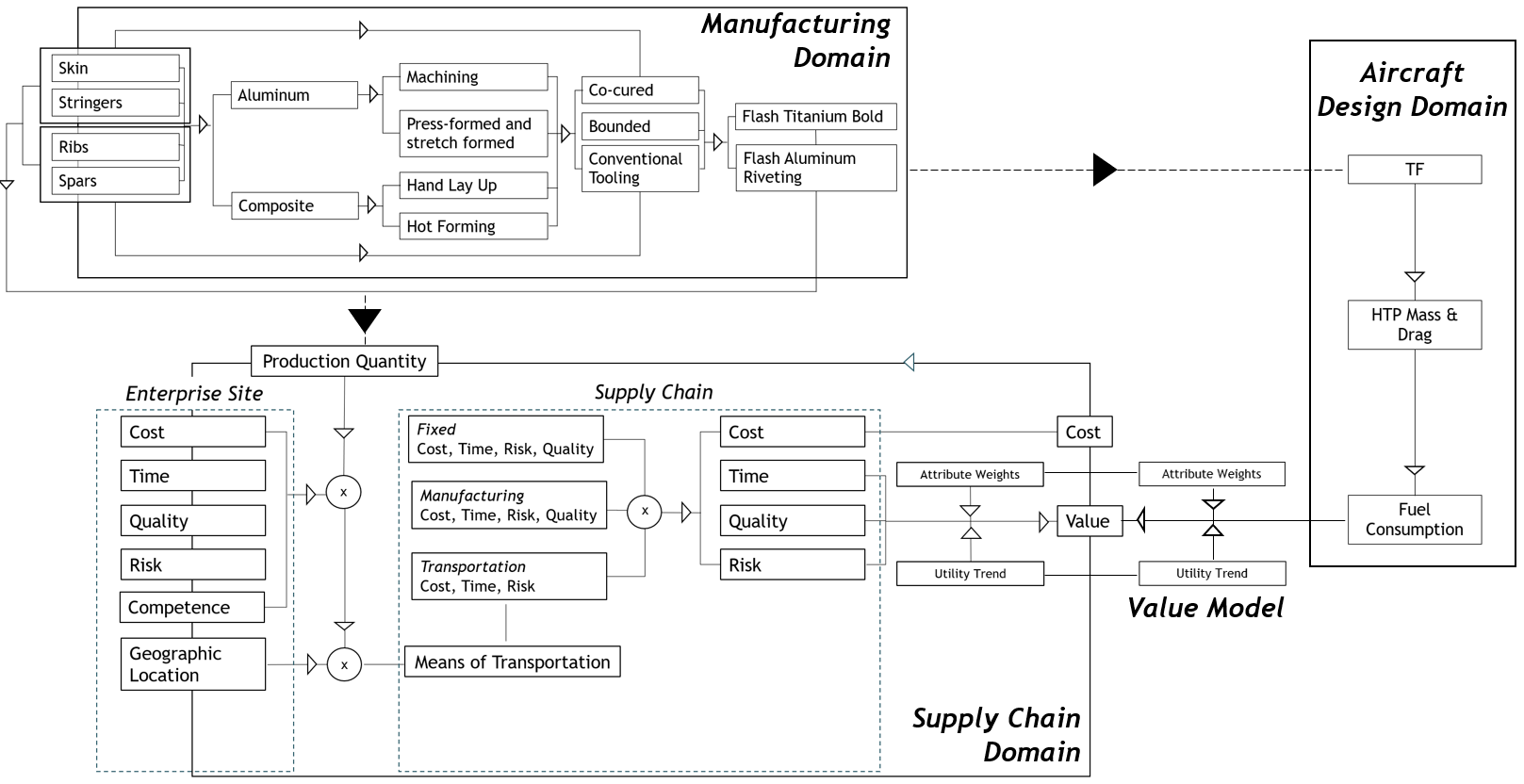}
  \caption{MDA including Manufacturing, Supply Chain and Aircraft Design Variables.}
  \label{fig:MDA_AC2}
\end{figure}

The objective is to have the ability to select the best HTP to produce by trading performance vs manufacturing costs vs supply chain performance.
In this study,  HTP structure breakdown  includes skin, stringers, spars and ribs that impact the three domains:
\begin{itemize}
     \item the manufacturing domain deals with the choices of materials, manufacturing and assembly processes feasible for the aircraft components. For each combination of materials,  a Technology Factor (TF), a dimensionless number ranging from 0 to 1, is produced.
     \item the  design domain deals with the estimation of the aircraft performance. The Technology Factor is considered in the evaluation as it  quantifies the impact that materials, manufacturing and assembly processes, have on the mass and drag of the HTP and, consequently, on the fuel consumption of the vehicle in cruise. The estimation of the fuel consumption is the main output of the design domain.
     \item the supply chain  domain estimates  the production performance  by characterizing the enterprises involved in the aircraft production and assembly. Here, each component of the HTP is characterized by a Production Quantity (PQ) that indicates the percentage of this component that each enterprise has to produce.  The cost, time, quality and  risk are the main outputs of this domain.
\end{itemize}     
For more details regarding this study, the reader can refer to~\cite{AGILE4_AVIATION_AC2:22,AGILE4_INCOSE_1:23,AGILE4_INCOSE_2:23}. Figure~\ref{fig:MDA_AC2} proposes a sketch of the interaction between the three domains to build the MDA.

In a first step, the focus was on the supply chain aspect in order to define the formalization, the settings of the optimizer, and the settings for remote calls. During this step,  the initial problems are used to calculate the full enumeration of solutions in order to compare the performance of the optimizer with the true Pareto Front of the database. 
Several increasingly complex problems were tackled, starting with a reduced number of components (for instance only skins and stringers), limited PQ numbers, and limited production sites. Eventually  an optimization problem considering all the HTP components and all possible supply chain sites was successfully tackled.

In a second step, the activities focus on integrating the manufacturing options in the overall optimization problem. It requires to add more variables, namely the choice of materials and their associated manufacturing processes for each HTP component. In addition, two constraints have now to be considered: 
\begin{itemize}
     \item the material incompatibility, that aims at discarding all the unfeasible combinations of material and manufacturing processes for the HTP components (for instance, an HTP can not be made using both skin in Aluminium - Machining and ribs in Thermoset - Hand Lay Up);
     \item the minimal supply chain site competence that checks that no 
 combination of skin, spar, stringer and ribs (depending of their material and manufacturing process) can be produced on a site with a competence of 0. 
\end{itemize}  
Table~\ref{tab:pb_ac2} presents the optimization problem in more details. Five objectives are considered, Cost, Risk, Time and Quality that are related to the supply chain domain and Fuel Consumption that is connected to the overall aircraft design and manufacturing domain. The variables represent all the possible choices of the HTP components characteristics and are treated as categorical variables in the optimization problem. Two types of variables are considered for each HTP component: 
\begin{itemize}
\item the choice of production and assembly site. Here the PQ have been frozen to 1, but all the possible production sites are available. Each component can therefore be produced among 21 different sites, either belonging to the OEM (Original Equipment Manufacturer) or to a supplier or sub-suppliers. 
\item the choice of materials and their associated manufacturing processes. Depending of the component, several combinations of material and manufacturing process can be selected (up to 6 for skin). For instance, choices could be Aluminium - Machining,  Thermoset - Hand Lay Up or Thermoset - Infusion Process.
\end{itemize}
Finally, the two constraints described earlier are included.


\begin{table}[!htb]
   \caption{Definition of the complete  problem.}
   \small
      \begin{tabular}{lclrr}
 & Function/variable & Nature & Quantity & Range\\
\hline
\hline
Minimize & Cost & cont & 1 &\\
Minimize &  Risk& cont & 1 &\\
Minimize & Time & cont & 1 &\\
Minimize &  Fuel Burn& cont & 1 &\\
Maximize & Quality & cont & 1 &\\
 & \multicolumn{2}{l}{\bf Total objectives} & {\bf 5} & \\
\hline
with respect to & skin prod. location & cat  &  21 levels & geographic sites\\
 & spar production location & cat  &  21 levels & geographic sites\\
  & stringer production location & cat  &  21 levels & geographic sites\\
   & rib production location & cat  &  21 levels & geographic sites\\
& skin   Material \& Manuf. process  & cat &  6 levels & Alu-Machining,...\\
& spar   Material \& Manuf. process  & cat  &  5 levels & Alu-Machining,...\\
& stringer   Material \& Manuf. process  & cat  &  4 levels & Alu-Machining,...\\
& rib  Material \& Manuf. process  & cat  &  5 levels & Alu-Machining,...\\
& \multicolumn{2}{l}{ Total categorical variables} & { 8} & \\ \cline{2-4}
 & \multicolumn{2}{l}{\bf Total relaxed variables} & {\bf 104} & \\
\hline
subject to & 
\multicolumn{2}{c}{ Material incompatibility} & 1 & \\
&\multicolumn{2}{c}{ Supply chain sites  }  & 1 & \\
 & \multicolumn{2}{l}{\bf Total constraints } & {\bf 2} & \\
\hline
\end{tabular}
   \label{tab:pb_ac2}
\end{table}

\begin{figure}[!htb]
   \begin{subfigure}[b]{.5\linewidth}
      \centering
      \includegraphics[width=\textwidth]{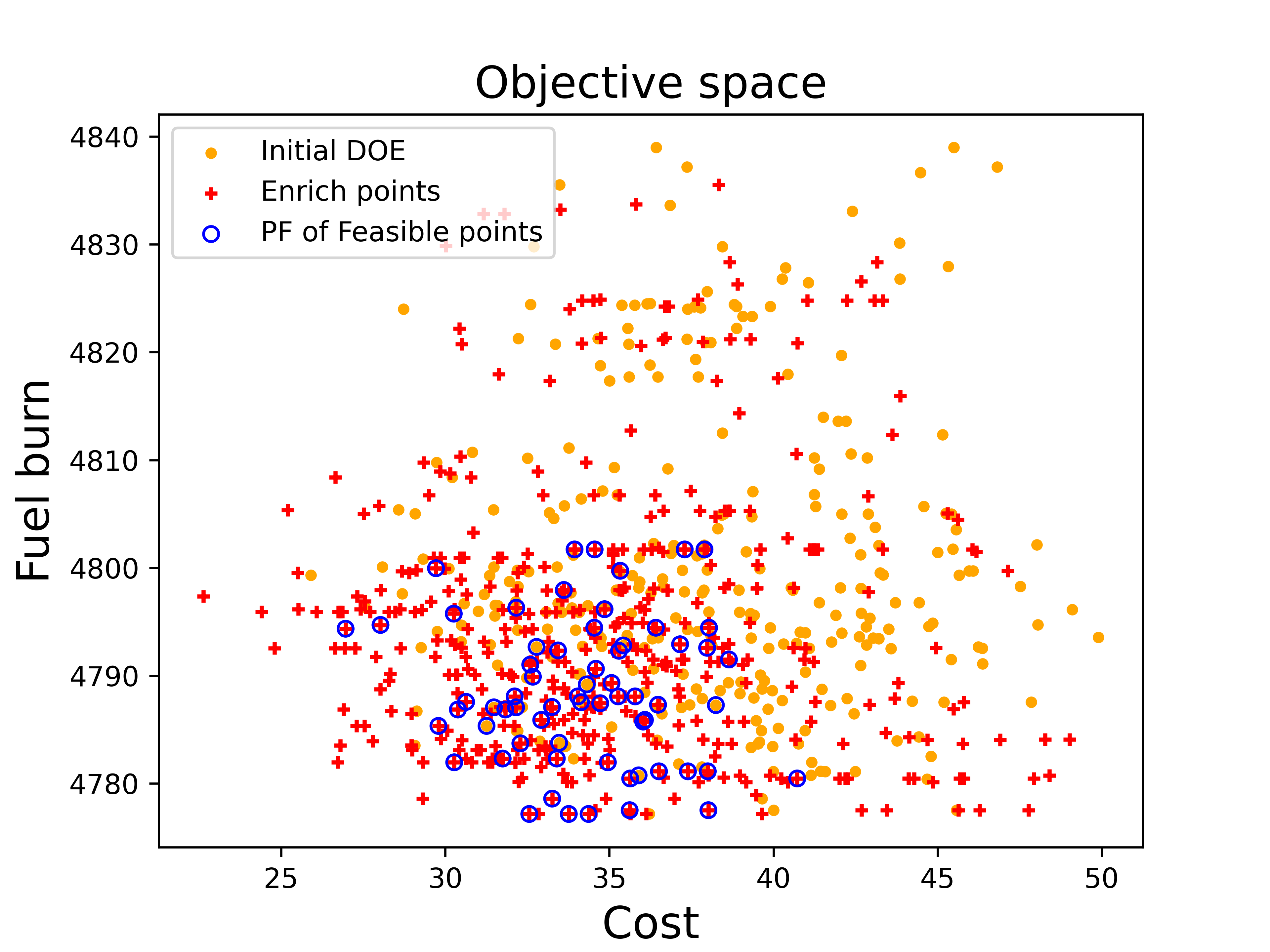}
      \caption{Fuel Burn vs Cost}
      \label{fig:myfig11}
      \end{subfigure}
      \begin{subfigure}[b]{.5\linewidth}
      \centering 
     {\includegraphics[width=\textwidth]{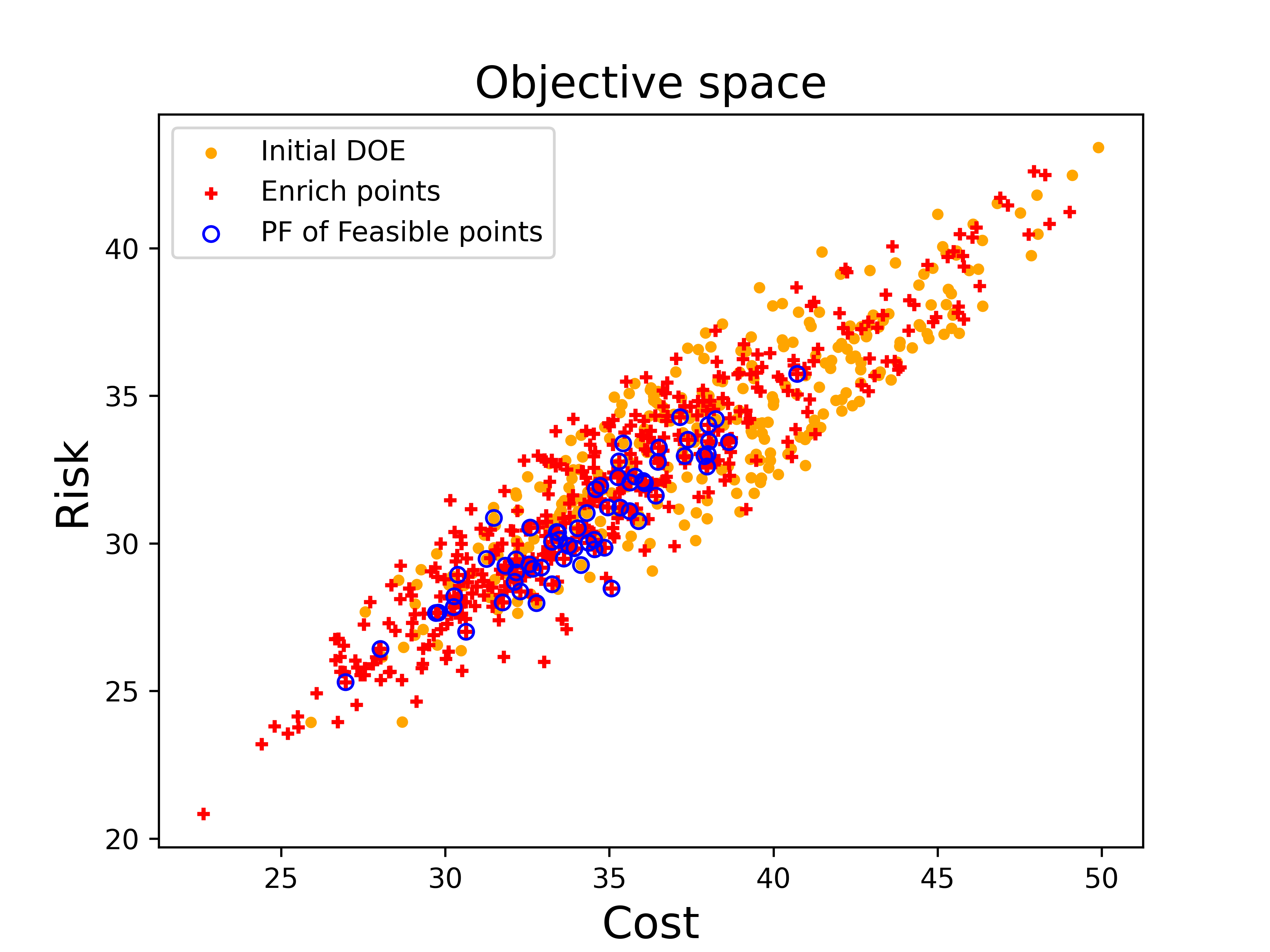}}
      \caption{Risk vs Cost}
      \label{fig:myfig12}
   \end{subfigure}
   \begin{subfigure}[b]{.5\linewidth}
      \centering 
      {\includegraphics[width=\textwidth]{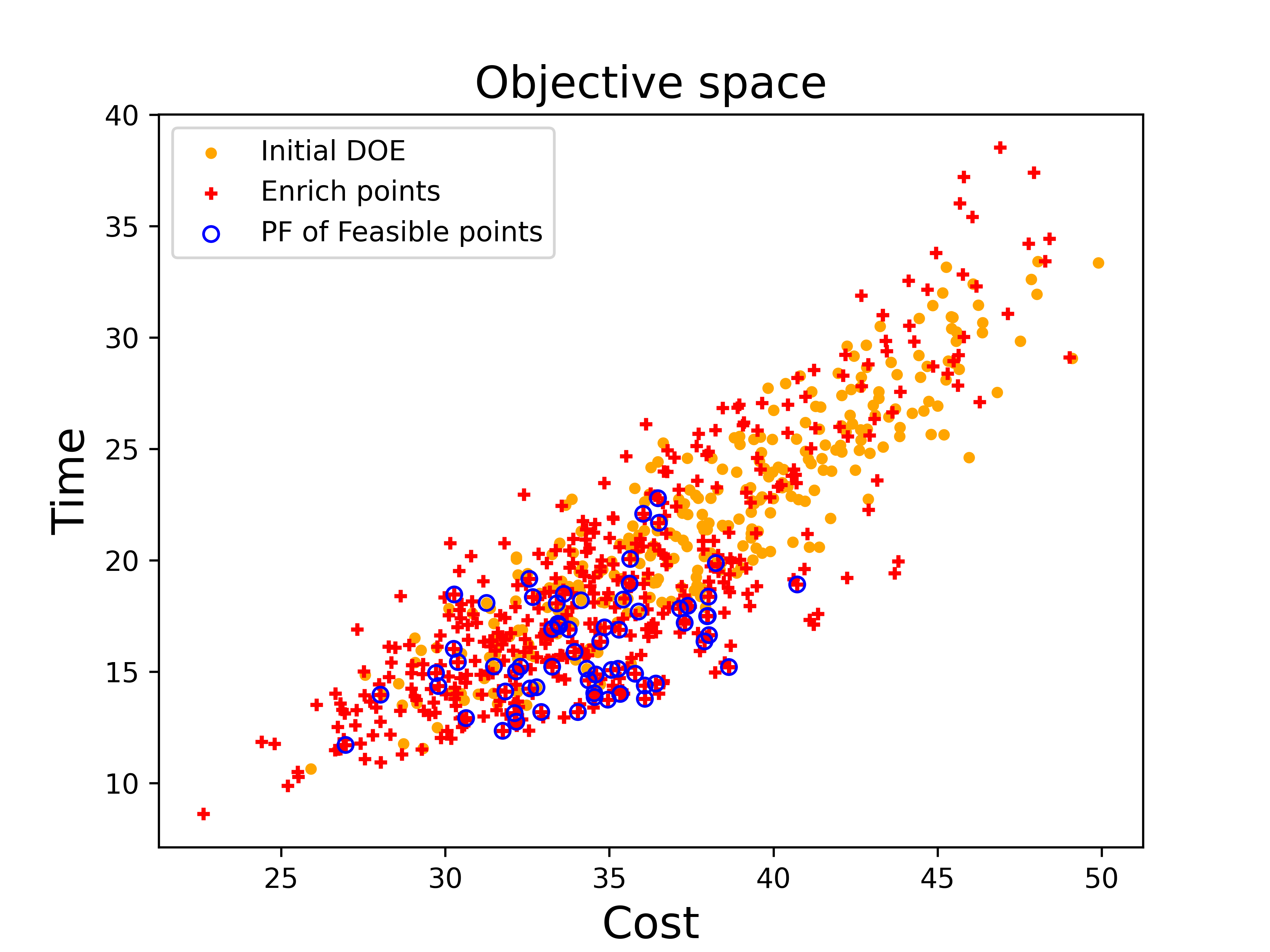}}
      \caption{Time vs Cost}
      \label{fig:myfig13}
   \end{subfigure}
      \begin{subfigure}[b]{.5\linewidth}
      \centering 
      {\includegraphics[width=\textwidth]{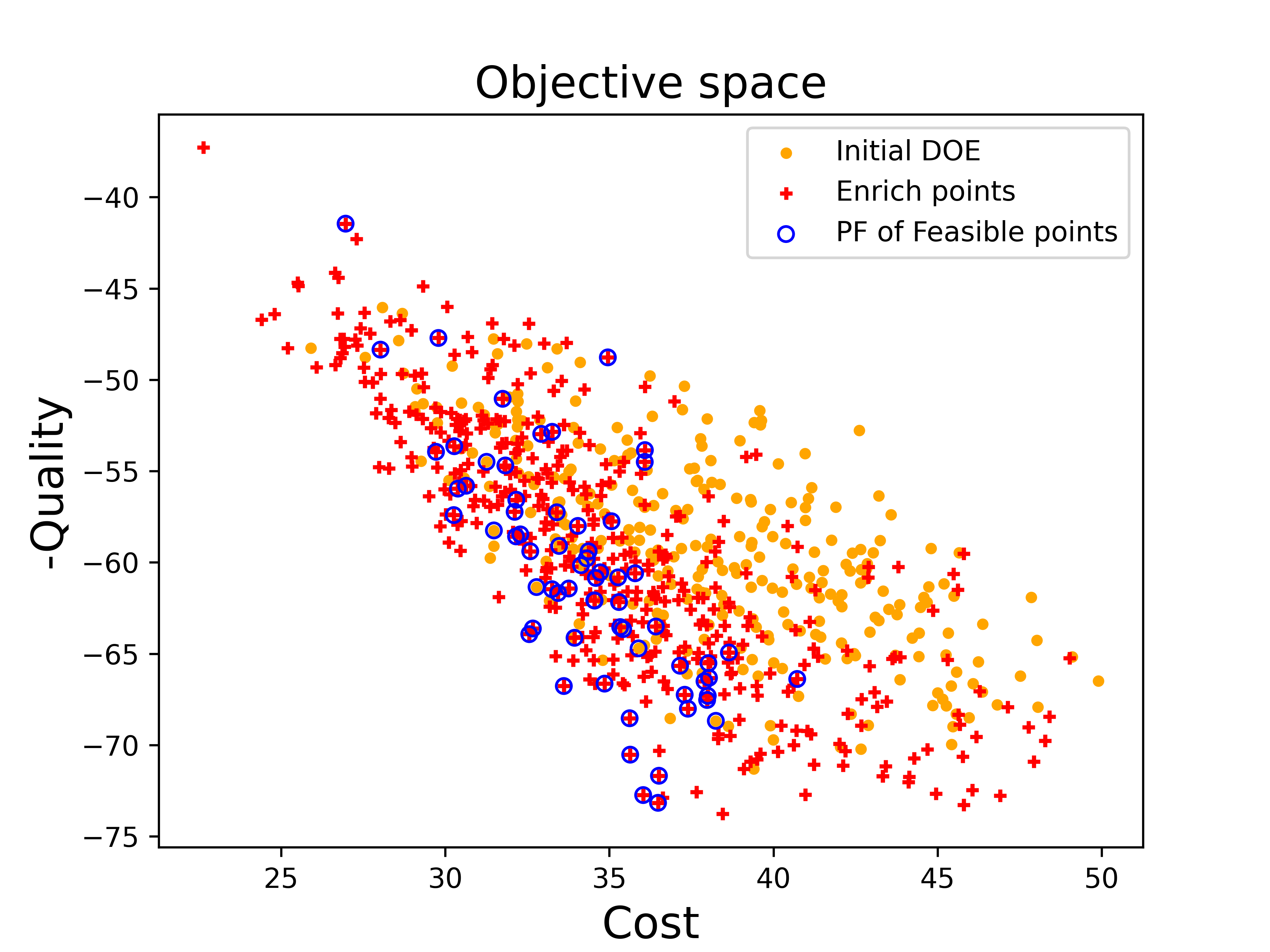}}
      \caption{- Quality vs Cost}
      \label{fig:myfig14}
   \end{subfigure}
   \caption{Database for five objectives: initial DOE of 300 points (orange dots), 450 enrichment points (red crosses) and,  from the complete database (750 points) 65 feasible points on the PF are identified by the blue circles.}
   \label{fig:AC2_PF}
\end{figure}

\begin{figure}[!htb]
   \begin{subfigure}[b]{.5\linewidth}
      \centering
      \includegraphics[width=\textwidth]{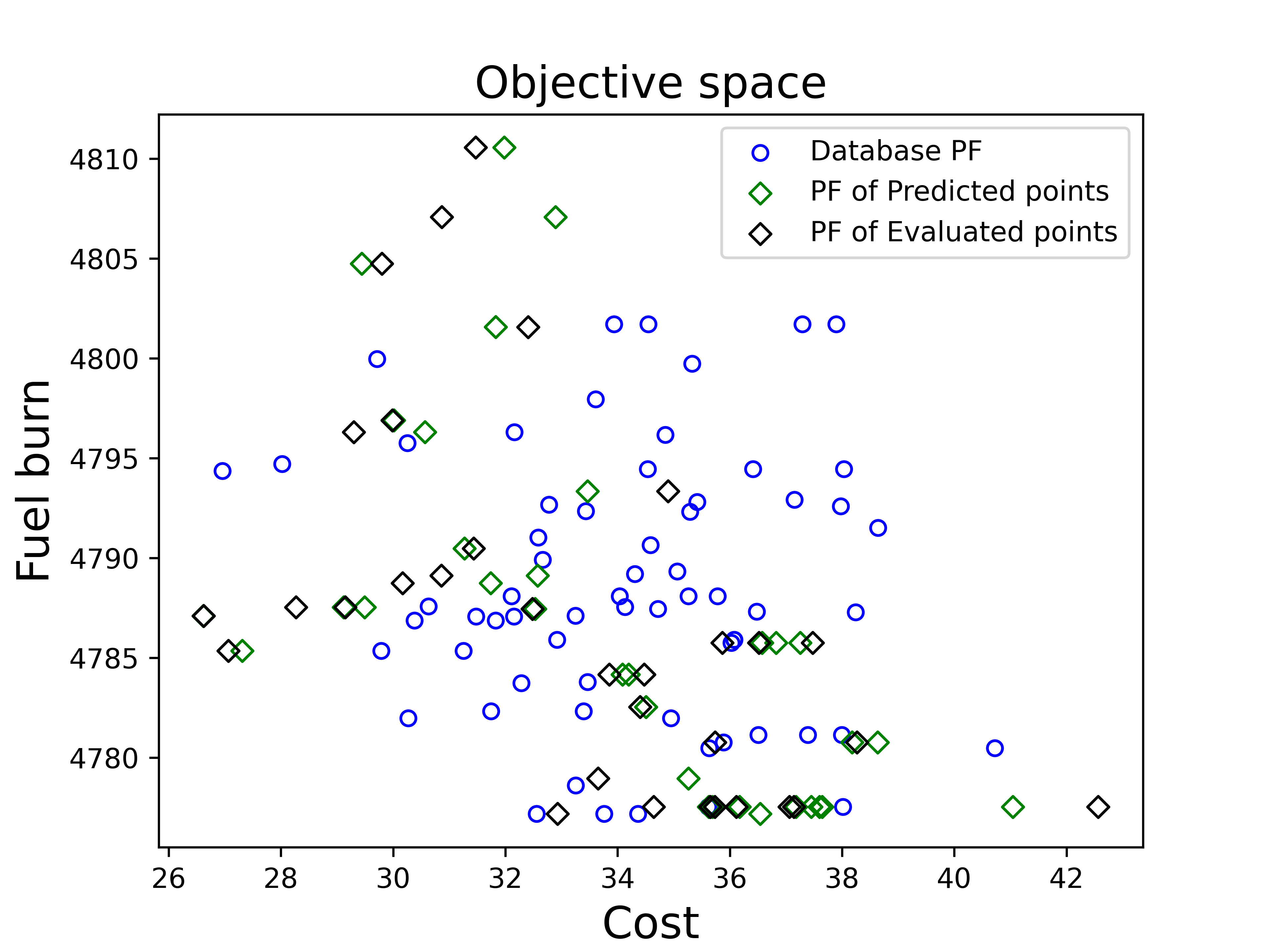}
      \caption{Fuel Burn vs Cost}
      \label{fig:myfig21}
      \end{subfigure}
      \begin{subfigure}[b]{.5\linewidth}
      \centering 
     {\includegraphics[width=\textwidth]{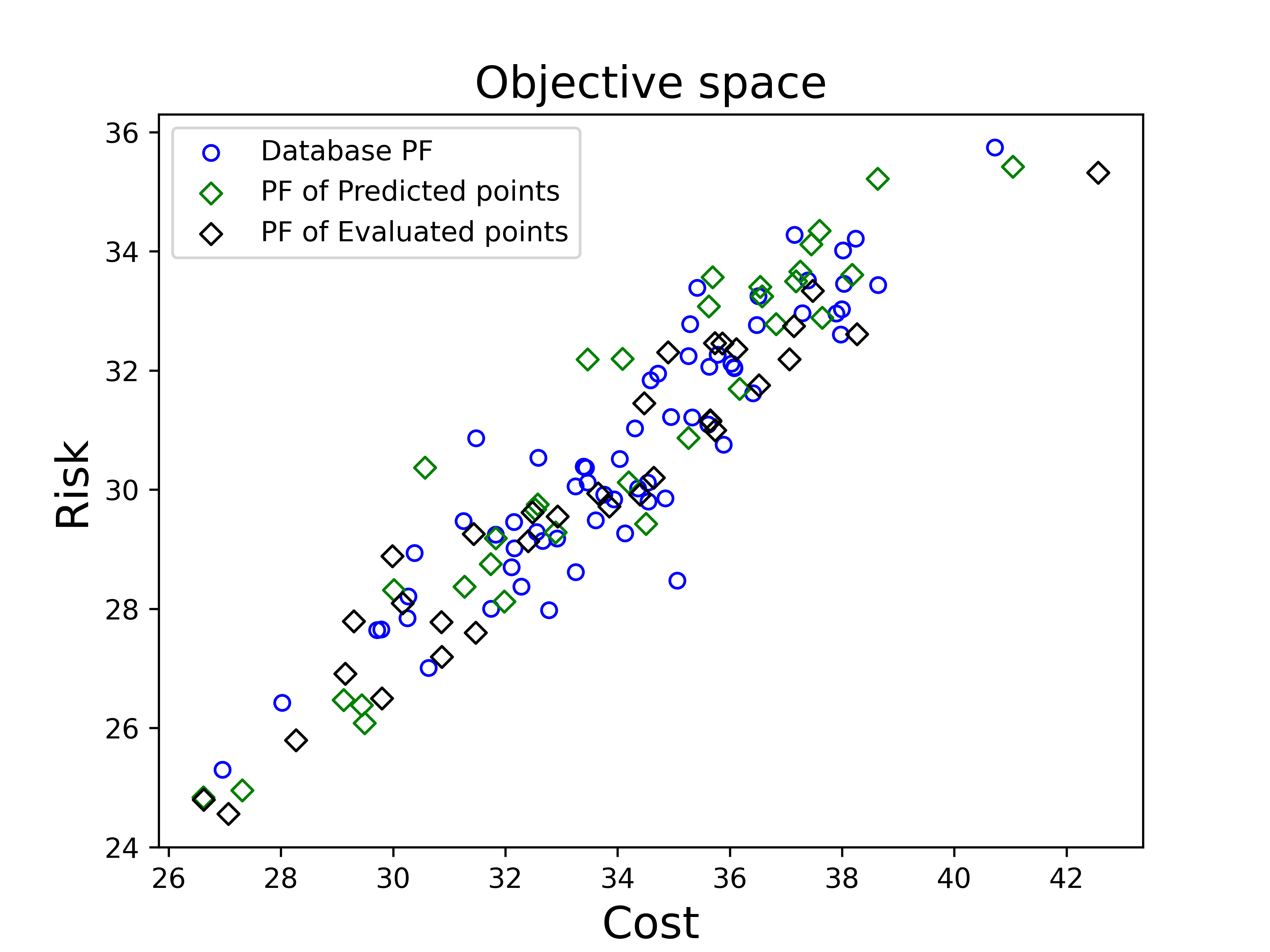}}
      \caption{Risk vs Cost}
      \label{fig:myfig22}
   \end{subfigure}
   \begin{subfigure}[b]{.5\linewidth}
      \centering 
      {\includegraphics[width=\textwidth]{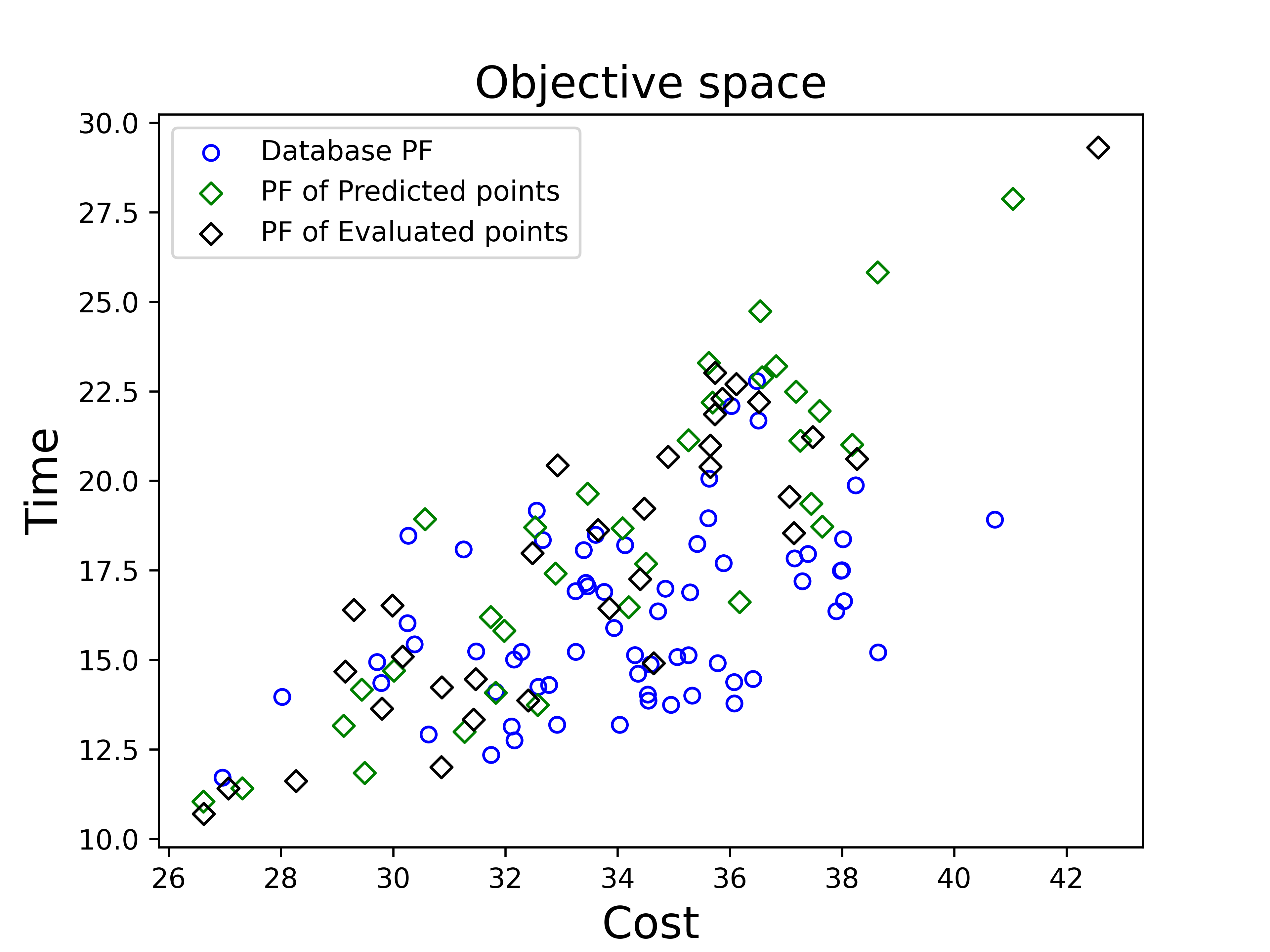}}
      \caption{Time vs Cost}
      \label{fig:myfig23}
   \end{subfigure}
      \begin{subfigure}[b]{.5\linewidth}
      \centering 
      {\includegraphics[width=\textwidth]{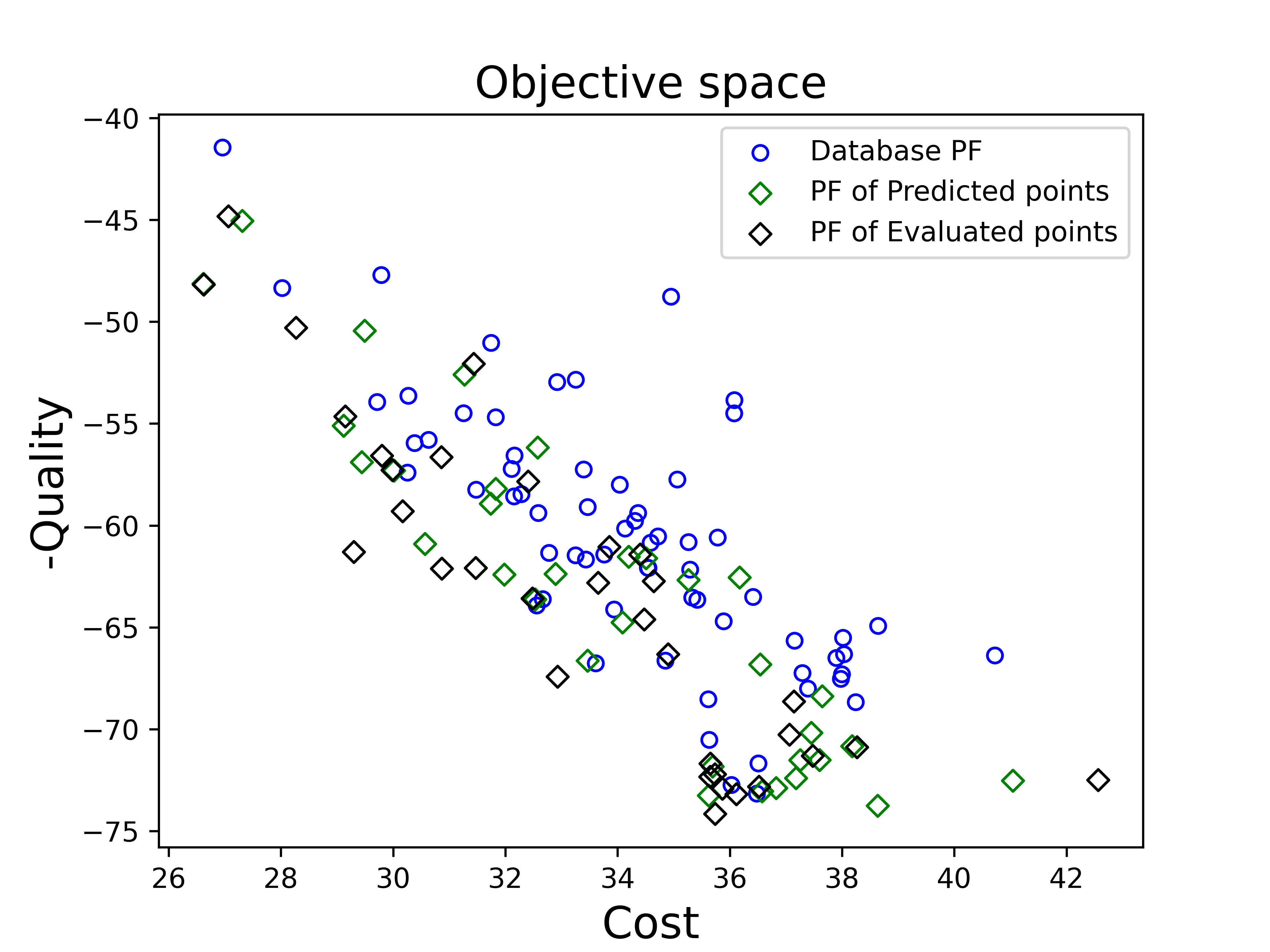}}
      \caption{- Quality vs Cost}
      \label{fig:myfig24}
   \end{subfigure}
   \caption{Comparison of PF (predicted and evaluated) with five objectives: 65 points from the PF database (blue circles), 33 points from the predicted PF (green diamonds) or evaluated with the workflow (black diamonds). The predicted and evaluated points are quite close to each other and complementary to the PF database.   }
   \label{fig:AC2_PF_END}
\end{figure}

Regarding optimization process, a 300-point DOE was executed to create the initial database of
design points for SEGOMOE and to verify the correct behavior of the MDO workflow.
Then, SEGOMOE was used to generate 450 additional infill points to explore the design
space and extend the Pareto front.
The main results of the optimization process are depicted in Figure~\ref{fig:AC2_PF} and Figure~\ref{fig:AC2_PF_END}. 
Figure~\ref{fig:AC2_PF} presents the PF database obtained at the end of the optimization run with 450 iterations. The initial DOE points are represented by orange dots and can provide information concerning the workflow behavior. The enrichment process of SEGOMOE added the 450 red crosses points that extended the database in direction of the Pareto Front (as it can be seen, for instance, 
 on the -Quality vs Cost Figure~\ref{fig:myfig14}). As the optimization is considering two constraints,  not all the database points are feasible and, at the end of the process,  65 points are identified on the PF (all highlighted with blue circles). 
 
 As indicated in Algorithm~\ref{algo:SEGOMOE_Multiobj}, a post-processing step is then performed applying NSGA-II algorithm to construct the predicted PF (using the GP models built upon the database). Figure~\ref{fig:AC2_PF_END} illustrates the results of this step with the PF database still depicted by blue circles. The predicted PF is represented by green diamonds, consisting of 33 candidates. It can be observed that both databases appear intermingled, indicating that the PF database already exhibits promising performance. Nevertheless,  as an additional step, all the points belonging to the predicted PF have been evaluated with the workflow and the outputs are depicted in black diamonds. Two pieces of information can be derived from this evaluated PF. Firstly, the evaluated candidates closely align with the predicted values (with the exception of a few), indicating the accuracy of the GP models, despite being built upon 750 points. Secondly, the PF database appears to maintain its good performance, characterized by its proximity to the evaluated PF. To verify this, both the PF database and evaluated PF were sorted, revealing that 52 points (out of 65) from the PF database and 26 points (out of 33) from the evaluated PF contribute to the final combined PF. As expected, the post-processing step of SEGOMOE successfully populates the PF identified by the algorithm, thereby enriching the available choices for the application stakeholders.

\section{Conclusions and perspectives}\label{sec:conclu}
Within the frame of AGILE 4.0 project, multi-objective problems involving mixed integer variables have been successfully solved using SBO and BO optimizers based on two associated frameworks JPAD Optimizer and SEGOMOE. In terms of function evaluations, the number of calls to the expensive black boxes (objectives and constraints) is still small compared to the number of design variables involved. SEGOMOE has been successfully applied to three realistic problems involving 2 to 5 objective functions with mixed integer variables. The use of continuous relaxation strategy to handle the categorical variables has been coupled with PLS to reduce the increase of dimension allowing to consider up to 104 choices.  

To apply BO to even more complex multidisciplinary processes, several developments need to be addressed. Future work will focus on extending the handling of hierarchical and mixed discrete variables while improving computational efficiency for larger databases. Specifically, with regards to mixed integer variables, the intention is to integrate recent advancements done within the SMT toolbox concerning mixed correlation kernels~\cite{Mixed_Paul} or hierarchical kernels~\cite{SMT2ArXiv} for Gaussian processes into BO and validate their effectiveness. These studies will be conducted within the context of the Horizon Europe COLOSSUS project~\footnote{https://colossus-sos-project.eu/}. The COLOSSUS project aims to develop a system-of-systems design methodology that facilitates the integrated optimization of aircraft, operations, and business models. This methodology will be applied to intermodal transport and wildfire-fighting scenarios.


\section*{Acknowledgements}
This work is part of the activities of ONERA - ISAE - ENAC joint research group.
The research presented in this paper has been performed within the framework of the AGILE 4.0 project (Towards Cyber-physical Collaborative Aircraft Development) and has received funding from the European Union Horizon 2020 Programme under grant agreement n${^\circ}$815122. Some methodology developments have been performed in the framework of the COLOSSUS project (Collaborative System of Systems Exploration of Aviation Products, Services and Business Models) and has received funding from the European Union Horizon Programme under grant agreement n${^\circ}$101097120.
The authors are also grateful to the partners of the AGILE 4.0 consortium for their contribution and feedback.



\bibliography{main}

\end{document}